\documentclass[12pt]{article}

\usepackage{booktabs}
\usepackage{tikz}
\usetikzlibrary{arrows.meta, positioning, shapes.geometric}
\usepackage{booktabs}
\usepackage{makecell}
\usepackage{siunitx}
\usepackage{threeparttable}
\usepackage{tabularx}
\usepackage{array}
\usepackage{scicite}
\usepackage{setspace}

\usepackage{caption}
\DeclareCaptionLabelFormat{bold}{\textbf{#1 #2}}
\usepackage{times}
\usepackage{subfiles}
\usepackage{amsmath}
\usepackage{graphicx} 
\usepackage{float}
\usepackage{rotating}
\usepackage{xurl}
\usepackage[colorlinks = true,
            linkcolor = blue,
            urlcolor  = blue,
            citecolor = blue,
            anchorcolor = blue]{hyperref}
\newenvironment{sciabstract}{%
\begin{quote} \bf}
{\end{quote}}

\newcounter{lastnote}

\title{Assessing the Carbon Emissions and Energy Consumption of U.S. Hyperscale Data Centers}

\author
{Gianluca Guidi,$^{1,2,3}$ Francesca Dominici,$^{1\dagger}$ Tiziano Squartini,$^{3}$ \\ Callaway Sprinkle, Jonathan Gilmour,$^{1}$ Kevin Butler,$^{4}$ Eric Bell,$^{5}$\\ Scott Delaney,$^{6}$ Falco J. Bargagli-Stoffi$^{1,7}$
\\
\footnotesize{$^{1}$Department of Biostatistics, Harvard T.H. Chan School of Public Health,}\\ \footnotesize{Boston, Massachusetts, USA
}\\
\footnotesize{$^{2}$Department of Computer Science, University of Pisa, Pisa, Italy}\\
\footnotesize{$^{3}$IMT School of Advanced Studies, Lucca, Italy}\\
\footnotesize{$^{4}$Environmental Systems Research Institute, Redlands, California, USA}\\
\footnotesize{$^{5}$Baxtel, Denver, Colorado, USA}\\ 
\footnotesize{$^{6}$Department of Environmental Health, Harvard T.H. Chan School of Public Health,}\\ 
\footnotesize{Boston, Massachusetts, USA}\\
\footnotesize{$^{7}$Department of Biostatistics, UCLA Fielding School of Public Health, Los Angeles, California, USA}\\
\scriptsize{$^\dagger$To whom correspondence should be addressed; E-mail:  fdominic@hsph.harvard.edu.}
}

\date{}

\begin{document} 

%\linenumbers
\baselineskip24pt


\maketitle 

\vspace{-1cm}

\singlespacing
\begin{sciabstract}
The rapid proliferation of hyperscale data centers (HDCs) in the US---mainly driven by the adoption of artificial intelligence---has raised concerns about this industry's environmental footprint. We compiled facility-level information on 403 US hyperscale data centers operating between May~2024 and April~2025 and estimated their electricity consumption, electricity sources, and attributable CO$_2$ emissions. Across different facility-load scenarios, these HDCs consumed approximately 68--99~TWh of electricity and were associated with about 37--54 million metric tons of CO$_2$. Under the central scenario, HDC electricity demand corresponded to approximately 1.8\% of total US electricity consumption, with roughly 54\% of attributed generation supplied by fossil-fuel sources. The HDC electricity-weighted average carbon intensity was approximately 545~gCO$_2$/kWh, about 48\% above the contemporaneous US national grid-average carbon intensity of 370~gCO$_2$/kWh. Our approach provides an attributional tool for assessing the environmental footprint of hyperscale data centers using the most recent EPA eGRID plant-level data.
\end{sciabstract}

{\noindent \textbf{One sentence summary:} Depending on facility-load assumptions, US hyperscale data centers consumed 68--99~TWh of electricity and were associated with 37--54~million metric tons of CO$_2$ between May~2024 and April~2025.}
\thispagestyle{empty}
\clearpage
\newpage

\pagenumbering{arabic}
\setcounter{page}{1}

\section*{Introduction}

\doublespacing

%{\color{blue} Overall comments: 1) Table 1: remove the .00 digits from columns 3, 6, 7. 2) Check that all references are up-to-date and correct (as I have been told that journals are implementing strict checks on references now).}

Data centers, facilities that house a large number of computing cores, serve as the backbone of modern digital infrastructure \cite {crawford2024generative,erdenesanaa2023ai,patterson2021carbon,mytton2022estimates,Shehabi2024DataCenter,Han2025}. They provide the critical resources needed to process, store and distribute large amounts of data, supporting applications such as cloud computing, artificial intelligence (AI), online services, e-commerce, social networks, and scientific research \cite{kamiya2024energy}. These facilities are inherently energy-intensive, and the primary electricity demands stem from computational power and cooling systems \cite{sun2021prototype}. Servers consume significant electricity by performing complex computations and generating substantial heat as a byproduct. To prevent overheating and ensure peak performance, stability, and longevity of the hardware, extensive cooling systems are employed, particularly in high-performance computing environments \cite{dayarathna2015data}.

\textit{Hyperscale data centers} (HDCs) are the largest and most powerful class of data centers. These facilities are designed to handle huge computing workloads and data storage requirements, supporting services such as cloud computing, large-language model training, and AI tool deployment on a global scale. HDCs differ significantly from traditional enterprise and colocation data centers in both size and power capacity. They typically consist of large modular units that span approximately 2,000 square meters each, allowing flexible expansion as demand grows \cite{Shehabi2024DataCenter}. These facilities have large power requirements, with average power capacities exceeding 40 megawatts, enough to power tens of thousands of servers simultaneously. In this study, we operationally define a hyperscale data center as a facility that satisfies three criteria: (i) it is operated by a hyperscale cloud/content provider or a colocation operator serving hyperscale tenants at scale; (ii) it has reported or inferred facility electrical capacity in the tens of megawatts; and (iii) its building or campus footprint is consistent with large-scale modular hyperscale architecture as verified through imagery and footprint data. We do not impose a single square-footage threshold because newer AI-oriented facilities can be substantially more power-dense than earlier cloud facilities.

In this paper, we study HDCs for two key reasons. First, HDCs have distinctive visibility and well-documented infrastructure architectures, allowing more statistically robust assessments. Second, they represent a growing market share, accounting for a significant portion of US data center load and driving substantial expansion in overall electricity consumption. Although in 2014, more than 60\% of server electricity consumption was in internal data centers, by 2023, this fell to nearly 10\%, with hyperscale facilities accounting for a large share of the sector's energy footprint \cite{Shehabi2024DataCenter}. This recent prominence stems primarily from the rise of cloud and AI workloads, especially after the commercial deployment of large-language-model applications, which has profoundly impacted demand for HDC compute. Cryptocurrency mining is a distinct electricity-intensive activity with different hardware and cooling characteristics, and we exclude it from the present analysis because it has been studied elsewhere\cite{guidi2025environmental}. Various contributions \cite{stokel-walker2024generative,strubell2019energy, Shehabi2024DataCenter} have reported that the number and size of data center facilities---including HDCs---have increased substantially. Shehabi et al. \cite{Shehabi2024DataCenter} have estimated that hyperscale and colocation data centers will account for over 90\% of server electricity consumption by 2028, exacerbating the strain on the electrical power grid.

The peer-reviewed literature presents a range of estimates of the electricity consumption of data centers \cite{mytton2022estimates}. Data center electricity use saw substantial increases in the early 2000s but remained nearly constant throughout the 2010s despite industry growth, mainly due to efficiency improvements such as better cooling systems, improved power management, increased server utilization, and reduced idle server power \cite{Shehabi2016DataCenter}. However, electricity demand has increased in recent years, mainly driven by hyperscale facilities that meet AI demands \cite{Shehabi2024DataCenter}. As electricity demand from data centers has increased, so have their emissions. These emissions depend critically on the power source or the energy mix used, with variations in the composition of the regional grid that significantly influence the calculations of the carbon footprint \cite{ferreira2018estimating}. Despite increased attention on US data center environmental impacts and growing numbers of institutional reports---see, e.g., \cite{eia_energyconsumption,Shehabi2024DataCenter,Shehabi2016DataCenter,EESI2025}---, peer-reviewed literature specifically examining hyperscale facilities' emissions is scarce.

An exception is Siddik et al.\cite{siddik2021environmental} who estimated that total greenhouse emissions from HDCs facilities were 10.5 million tons in 2018. However, these estimates are now outdated given recent increases in electricity demand driven by Bitcoin mining and AI. Similarly to what we do, Siddik et al. rely on private providers for data center information and implemented a bottom-up approach to estimate the electricity use based on data center floor space. However, to our knowledge, their data pipeline was not further validated via satellite imagery. More recently, Xiao et al. \cite{xiao2025environmental} projected that AI servers alone could generate 24--44 million tons CO$_{2}$ in the US between 2024 and 2030, highlighting the urgency of understanding current facility-level emissions patterns.

In this paper, we scale the work by \cite{siddik2021environmental} by introducing a data pipeline that harmonizes heterogeneous data sources and validates (through satellite imagery) facility-level data on 403 HDCs operating from May 2024 to April 2025 (more details of the validation are provided in the Supplementary Materials, see also Fig.~\ref{fig:data_validation_pipeline}). We excluded cryptocurrency mining data centers as they have been extensively studied elsewhere---see, e.g. \cite{guidi2025environmental,stoll2019carbon}.  

For HDCs, we: 1) identified their electricity consumption; 2) pinpointed the power plants that supply the electricity to that HDCs; 3) identified each individual power plant's share of electricity supplied and the fuel used to produce the electricity; and 4) estimated the aggregate-level CO$_{2}$ emissions, by state and balancing authority. We estimated and compared CO$_{2}$ emissions and \textit{carbon intensities}---defined as the amount of CO$_{2}$ produced per unit of electricity consumed---for HDCs versus other relevant sectors of the US economy. Finally, we developed a public-facing web platform to track CO$_{2}$ emissions, electricity load estimates, and total electricity demand from HDCs in the balancing authority region and at the state level.  

\section*{Results}

\subsection*{Hyperscale Data Centers and their Electricity Consumption}

\begin{figure}
\begin{center}
\includegraphics[width=1\textwidth]{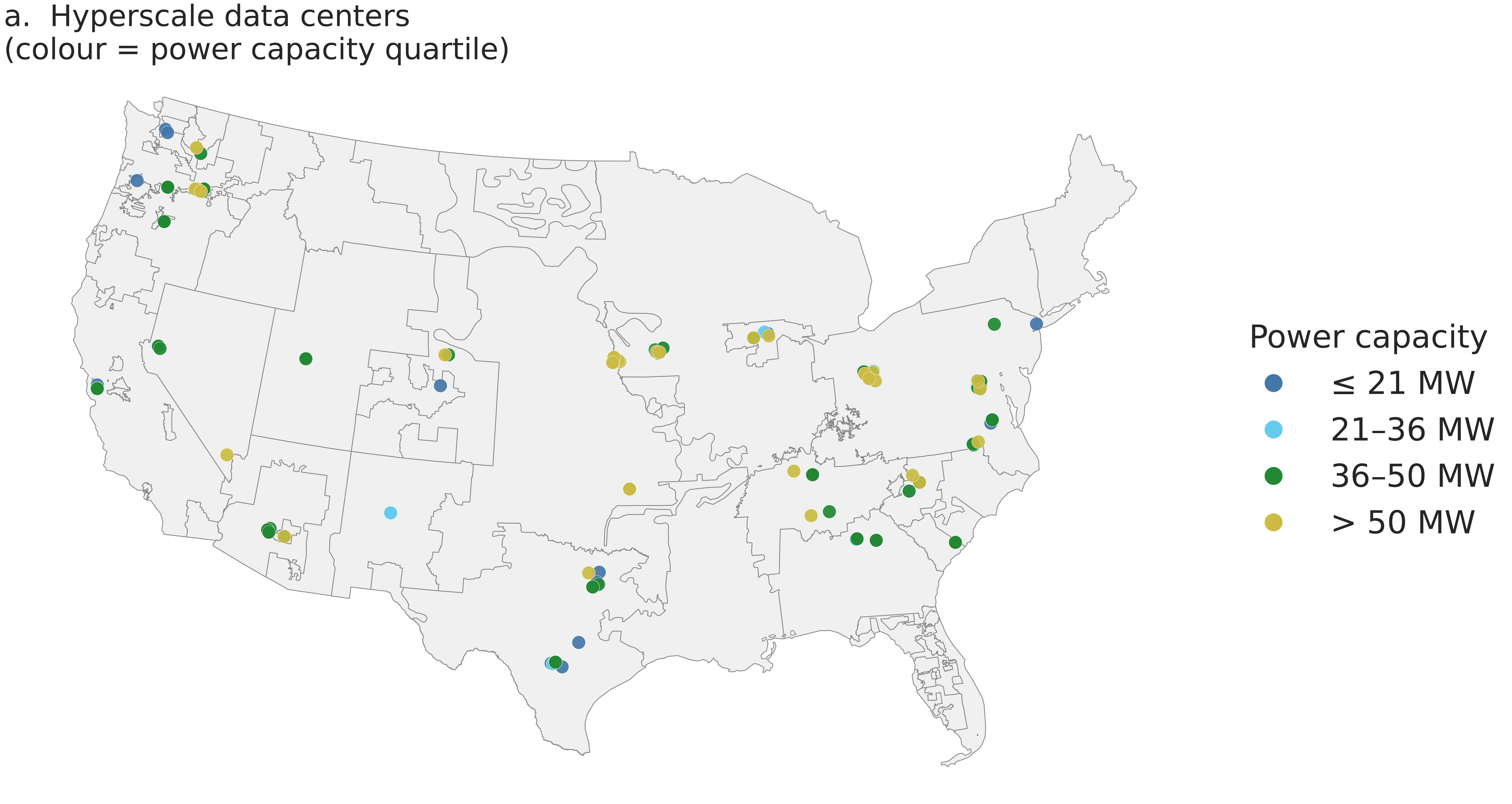}
\includegraphics[width=1\textwidth]{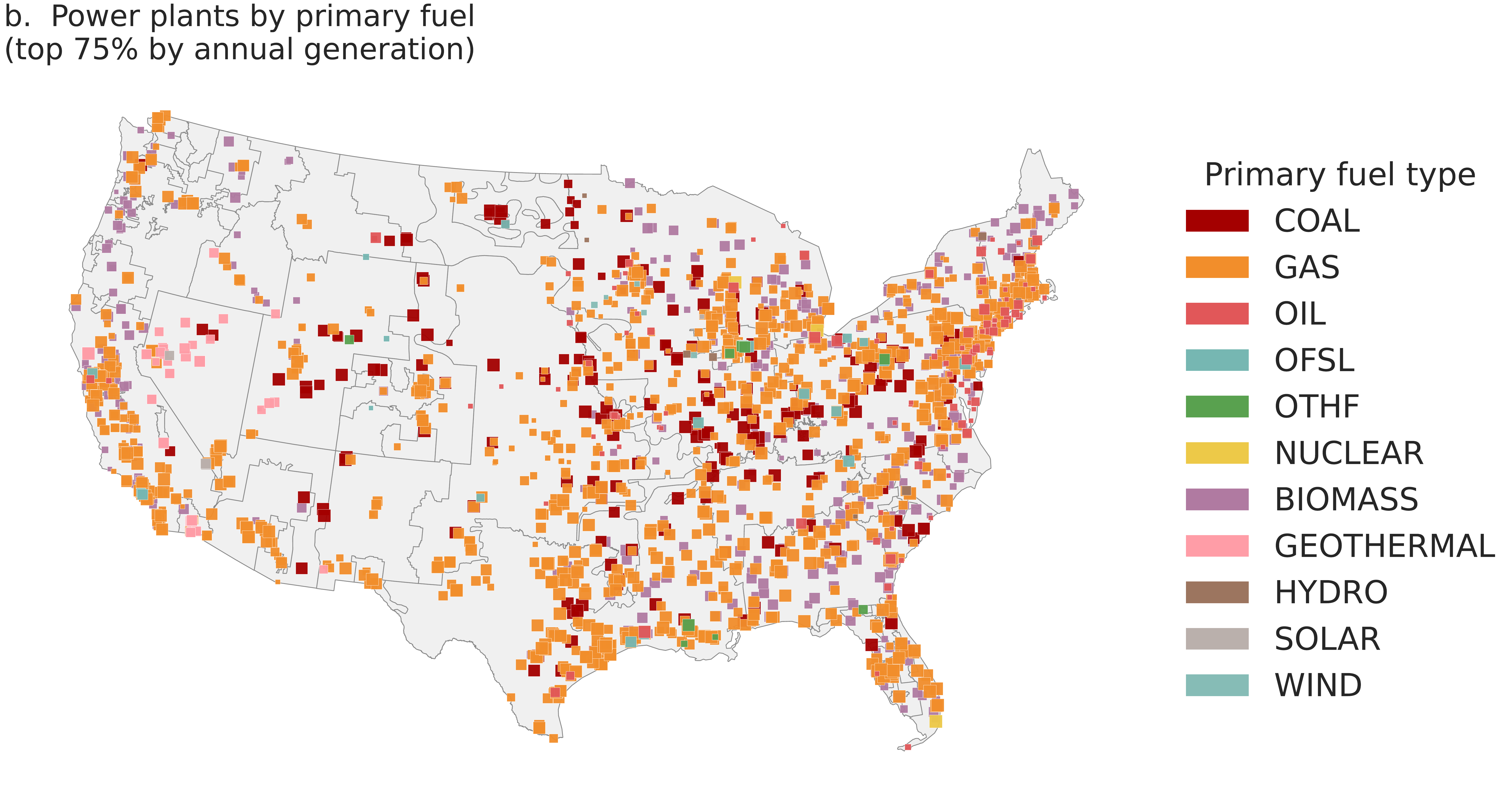}
\end{center}
\caption{\textbf{Geographic distribution of hyperscale data centers (1a) and power plants in the contiguous US (1b), overlaid with balancing authority regions.} 
Figure~1a shows the hyperscale data centers in the analytical dataset. Hyperscale data centers are divided into quartiles based on their power capacity.
Many HDCs are geographically clustered, so dots sit on top of each other, visually collapsing into a smaller number of blobs.
Figure~1b displays the subset of operational power plants shown for visual clarity after excluding the bottom 25\% of plants by annual net electricity generation. Power plants are colored by primary fuel type, and marker sizes are proportional to annual electricity generation. The plant-level attribution calculations use EPA eGRID2023 generation, emissions, and fuel-mix data within each balancing authority, consistent with the attributional framework used in the analysis.}
\label{fig:location}
\end{figure}

%\begin{figure}
%\begin{center}
%\includegraphics[width=1\textwidth]{Images/dc_plants.pdf}
%\includegraphics[width=0.48\textwidth]{Images/Figure 1.pdf}
%\end{center}
%\caption{\textbf{Geographic distribution of hyperscale data centers and power plants in the contiguous US, overlaid with balancing authority regions.} This figure shows the 403 hyperscale data centers and 4,819 operational power plants included in our analysis for the study period from May 2024 to April 2025. The map is displayed at the balancing authority (BA) level, representing regions where electricity supply and demand are managed in real time. The size of each hyperscale data center marker is proportional to its power capacity, while power plants are colored by their primary fuel type.
%}
%\label{fig:location}
%\end{figure}

We identified 403 HDCs located throughout the contiguous US that operated from May 2024 to April 2025 (Fig. \ref{fig:location}). Virginia had more HDCs (142) than any other state, followed by Oregon (56) and Ohio (38). 
We estimate that the 403 data centers in our study consumed approximately
68--99~TWh of electricity during the study period, depending on the
facility-load scenario. The central scenario ($u=0.58$, informed by
bottom-up power-flow modeling and an independent PUE-implied check) gives
roughly 82~TWh; lower and higher scenarios ($u \in \{0.48, 0.663, 0.70\}$)
yield approximately 68, 93, and 99~TWh respectively. These values correspond
to roughly 1.5--2.2\% of total US electricity consumption. Unless otherwise
noted, maps and state-level tables in the main text show the central
scenario, while the full scenario range is reported in the Supplementary
Materials.

Based on benchmarks derived from the International Energy Agency's \emph{Energy and AI} report, which estimates total US data center electricity consumption at approximately 200~TWh in 2024 and implies a hyperscale share of roughly 120~TWh, our facility sample represents approximately 56--82\% of implied US hyperscale electricity consumption, depending on the facility-load scenario adopted. This comparison suggests our dataset is broadly consistent in scale with IEA's implied national hyperscale electricity, recognizing that both estimates carry substantial uncertainty: ours from facility-load assumptions and incomplete coverage, theirs from the methodology used to disaggregate hyperscale from total data center load \cite{iea2025energyai}. We therefore interpret our totals not as a single point estimate but as a range across four facility-load scenarios (Section~\ref{sm:s3_utilization}), which together bracket the uncertainty arising from how reported facility nameplate capacity translates into average operational electricity demand. Incomplete facility coverage acts as an additional source of uncertainty whose direction is not symmetric with the scenario range.

Our estimate exceeds the total annual electricity consumption of many individual states in the US \cite{EIA_fuel_use}. For comparison, the three highest electricity-consuming states — Texas, California, and Florida — consumed approximately 475, 251, and 248~TWh respectively. Meanwhile, states such as Indiana, Michigan, and Tennessee, each with annual total-sector consumption of approximately 100~TWh, have electricity footprints comparable to the total HDC consumption in our dataset~\cite{EIA_fuel_use}. Under the central scenario, total electricity consumption attributable to our 403 HDCs is approximately 3.6 times the 22.85~TWh reported by Siddik et al.~\cite{siddik2021environmental} for HDCs in 2018; across the scenario range, the ratio is approximately three- to four-fold, consistent with the continued displacement of smaller, less efficient data centers by hyperscale facilities.

From May~2024 to April~2025, the average facility power capacity of the HDCs in our sample was 39.9~MW, with a median of 36~MW. Here, facility power capacity refers to the total facility electrical capacity, defined as the maximum deliverable load at the facility’s electrical infrastructure, including IT load as well as cooling and electrical overhead. The average power density, defined as facility power capacity divided by total building floor area, was approximately 1{,}600~W/m$^{2}$. This figure is facility-inclusive (i.e., the denominator is total facility floor area, not IT-equipment-only area) and falls within the range of facility-inclusive hyperscale power densities reported by the Uptime Institute Global Data Center Survey~\cite{uptime2024globalsurvey}, Lei \& Masanet~\cite{lei2022climatepue}, and recent industry-analyst benchmarks~\cite{goldman2023ai}.We use this comparison as a broad plausibility check on the scale of the estimated capacities and not as a stand-alone validation of the downstream electricity and emissions calculations. Across utilization scenarios, the HDCs located in Virginia had the highest aggregate electricity consumption among all US states, followed by Oregon, Ohio, and Iowa. Under the central scenario, Virginia accounted for approximately 21~TWh, or roughly one quarter of total HDC electricity consumption in our dataset. Oregon followed with about 11~TWh ($\sim$13\%), Ohio with about 9~TWh ($\sim$10\%), and Iowa with about 7~TWh ($\sim$9\%). Together, these four states accounted for more than 50\% of total HDC electricity consumption in our analytical sample, and — given that our sample covers an estimated 56--82\% of the national hyperscale electricity footprint — likely represent a similarly dominant share at the national level.

\subsection*{Identifying and Characterizing Power Plants Supplying Electricity to Hyperscale Data Centers}

We identified 4,819 power-plant records represented in the attributional supply mix for the balancing authorities containing the 403 HDCs. The identification of these power plants was performed by applying an attribution method\cite{brander_most_2022}. We geolocated each HDC to its corresponding balancing authority (that is, the entity charged with maintaining grid reliability by matching generation and demand within a defined territory \cite{epa_egrid_faq}, see  Figure \ref{fig:location}). Under the attributional method, we assume that within each balancing authority,  the total HDCs electricity demand is distributed across power plants in proportion to their annual electricity generation. For example, a plant contributing 10\% of a region’s electricity is assumed to supply 10\% of the HDCs load in that region. The rationale for this attributional method is detailed in the Materials and Methods Section as well as the Supplementary Materials.

We find that 53.9\% of the electricity attributed to HDCs is supplied by fossil-fuel generation, 20.9\% by nuclear generation, and 25.3\% by renewable generation (Fig.~\ref{fig:fuel_mix}). These shares are computed as generation-share-weighted fuel shares within each balancing authority and then weighted by HDC electricity demand; they are not plant-count shares. Within the fossil component, natural gas contributes 38.2\% of total attributed electricity and coal contributes 15.2\%. Among non-fossil sources, nuclear contributes 20.9\%, while hydro, wind, solar, biomass, and geothermal together account for 25.3\% (see  Figure \ref{fig:location}).

\subsection*{Carbon Emissions of the Supplying Power Plants}

\begin{figure}
\begin{center}
\includegraphics[width=1\textwidth]{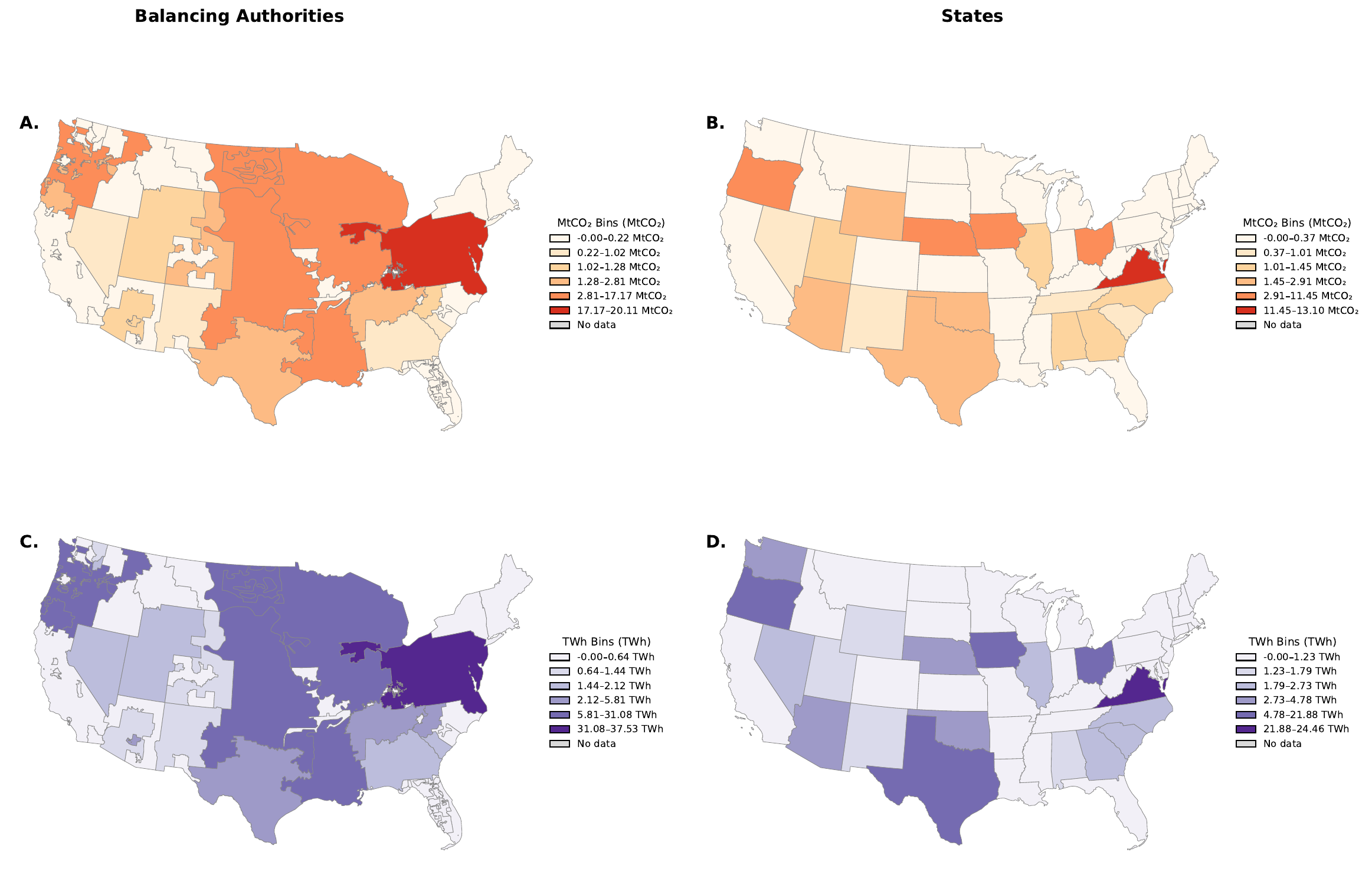}
\end{center}
\caption{\textbf{Hyperscale Data Center electricity consumption and CO$_{2}$
emissions.} (\textbf{a, c}) The balancing authority region in which a
hyperscale data center is located determines the mix of power plants that
supply its electricity and thus its attributable emissions. See
Fig.~\ref{fig:BA} for balancing authority regions and corresponding names.
(\textbf{b, d}) Maps at the state level show electricity consumption and
emissions for which the hyperscale data centers within the state are
responsible. Color bins represent percentile-based ranges: 0--20\%,
20--40\%, 40--60\%, 60--80\%, 80--99\%, and 99--100\%. Maps display the
central facility-load scenario ($u = 0.58$); the full scenario range is
reported in Supplementary Table~\ref{tab:state_scenarios}.}
\label{fig:powerload_emissions}
\end{figure}

After identifying the power plants represented in each balancing authority, we estimated the corresponding CO$_2$ emissions attributable to HDC electricity consumption using EPA eGRID2023 plant-level generation and emissions factors. Under the central facility-load scenario, total CO$_2$ emissions attributable to the 403 HDCs were approximately 45~Mt. Across the four facility-load scenarios, emissions ranged from approximately 37 to 54~Mt, roughly 3.5- to 5-fold the 10.5~Mt reported for HDCs in 2018 by Siddik et al.~\cite{siddik2021environmental}. The highest state-level attributable emissions occurred in Virginia, Oregon, Ohio, and Iowa (approximately 11.5, 4.9, 4.7, and 4.6~Mt respectively under the central scenario; see Table~1). Figure~\ref{fig:powerload_emissions} shows the distribution of electricity consumption and attributable emissions at the balancing-authority and state levels.

%\DIFaddbeginFL
\begin{table}[htbp]
\centering
\small
\setlength{\tabcolsep}{5.0pt}
\renewcommand{\arraystretch}{1.18}
\begin{threeparttable}
\caption{\textbf{Top ten states by CO$_2$ emissions attributable to hyperscale data center electricity consumption under the central facility-load scenario.}}
\begin{tabular}{
  l
  S[table-format=2.1, round-precision=1]
  S[table-format=2.1, round-precision=1]
  S[table-format=3.0, round-precision=0]
  S[table-format=1.2, round-precision=2]
  S[table-format=1.2, round-precision=2]
  S[table-format=4.0, round-precision=0]
  S[table-format=2.0, round-precision=0]
}
\toprule
& \multicolumn{3}{c}{\textbf{State totals}}
& \multicolumn{4}{c}{\textbf{Hyperscale data center averages per state}} \\
\cline{2-4}\cline{5-8}\noalign{\vskip 2pt}
\textbf{State}
& {\makecell{\textbf{Annual CO$_2$}\\\textbf{emissions}\\{\normalfont (Mt)}}}
& {\makecell{\textbf{Annual electricity}\\\textbf{use}\\{\normalfont (TWh)}}}
& {\makecell{\textbf{Number of}\\\textbf{HDCs}\\{\normalfont (\#)}}}
& {\makecell{\textbf{Electricity}\\\textbf{demand}\\{\normalfont (TWh/HDC)}}}
& {\makecell{\textbf{CO$_2$}\\\textbf{emissions}\\{\normalfont (Mt/HDC)}}}
& {\makecell{\textbf{Carbon}\\\textbf{intensity}\\{\normalfont (gCO$_2$/kWh)}}}
& {\makecell{\textbf{Facility size}\\{\normalfont (10$^3$ m$^2$)}}} \\
\midrule 
Virginia  & 11.5 & 21.0 & 142 & 0.15 & 0.08 &  535 & 18 \\
Oregon    &  4.9 & 11.0 &  56 & 0.20 & 0.09 &  440 & 23 \\
Ohio      &  4.7 &  9.0 &  38 & 0.23 & 0.12 &  535 & 25 \\
Iowa      &  4.6 &  7.0 &  26 & 0.27 & 0.18 &  645 & 42 \\
Nebraska  &  2.6 &  3.5 &   8 & 0.44 & 0.32 &  740 & 51 \\
Texas     &  2.4 &  4.6 &  22 & 0.21 & 0.11 &  535 & 30 \\
Oklahoma  &  2.1 &  2.9 &   8 & 0.36 & 0.27 &  740 & 44 \\
Arizona   &  1.9 &  3.2 &  12 & 0.27 & 0.16 &  595 & 40 \\
Wyoming   &  1.3 &  1.2 &   2 & 0.61 & 0.64 & 1035 & 18 \\
Illinois  &  1.3 &  2.3 &   8 & 0.29 & 0.16 &  535 & 35 \\
\bottomrule
\end{tabular}
\begin{tablenotes}[flushleft]
\footnotesize
\item Notes: State totals are reported under the central facility-load scenario ($u=0.58$), chosen because it is consistent with bottom-up power-flow estimates and with our independent PUE-implied check, using EPA eGRID2023 plant-level emissions and generation data. State totals (columns 2--3) are rounded to one decimal place; per-HDC averages (columns 5--6) to two decimal places; carbon intensity and facility size to the nearest integer. Per-state averages are computed across hyperscale data centers located in each state. Attributable emissions reflect the balancing-authority generation mix supplying data center electricity and therefore do not necessarily occur within the same state boundaries. Full scenario ranges by state are reported in Supplementary Table~\ref{tab:state_scenarios}.
\end{tablenotes}
\end{threeparttable}
\label{tab:table1}
\end{table}
%\DIFaddendFL

\subsection*{Carbon Intensity and Fuel Mix of the Supplying Power Plants}

Next, we evaluated the carbon intensity of the electricity attributed to HDCs. Approximately 90\% of HDC electricity demand is located in balancing authorities with carbon intensity above the US national average of about 370~gCO$_2$/kWh in 2023~\cite{ourworldindata_carbon_intensity}, the calendar year of grid generation underlying our eGRID2023 attribution layer. The HDC electricity-weighted average carbon intensity was approximately 545~gCO$_2$/kWh, about 48\% above the contemporaneous US national grid-average carbon intensity ($\sim$370~gCO$_2$/kWh in 2023). This indicates that HDCs in our sample are disproportionately concentrated in balancing authorities whose electricity supply is more carbon-intensive than the national grid average.

Virginia's dominant balancing authority, PJM, is the balancing authority with the highest HDC electricity demand in our dataset. Under eGRID2023, PJM's attributed carbon intensity is approximately 535~gCO$_2$/kWh, above the US national average \cite{ourworldindata_carbon_intensity}.

\begin{figure}
\begin{center}
\includegraphics[width=1\textwidth]{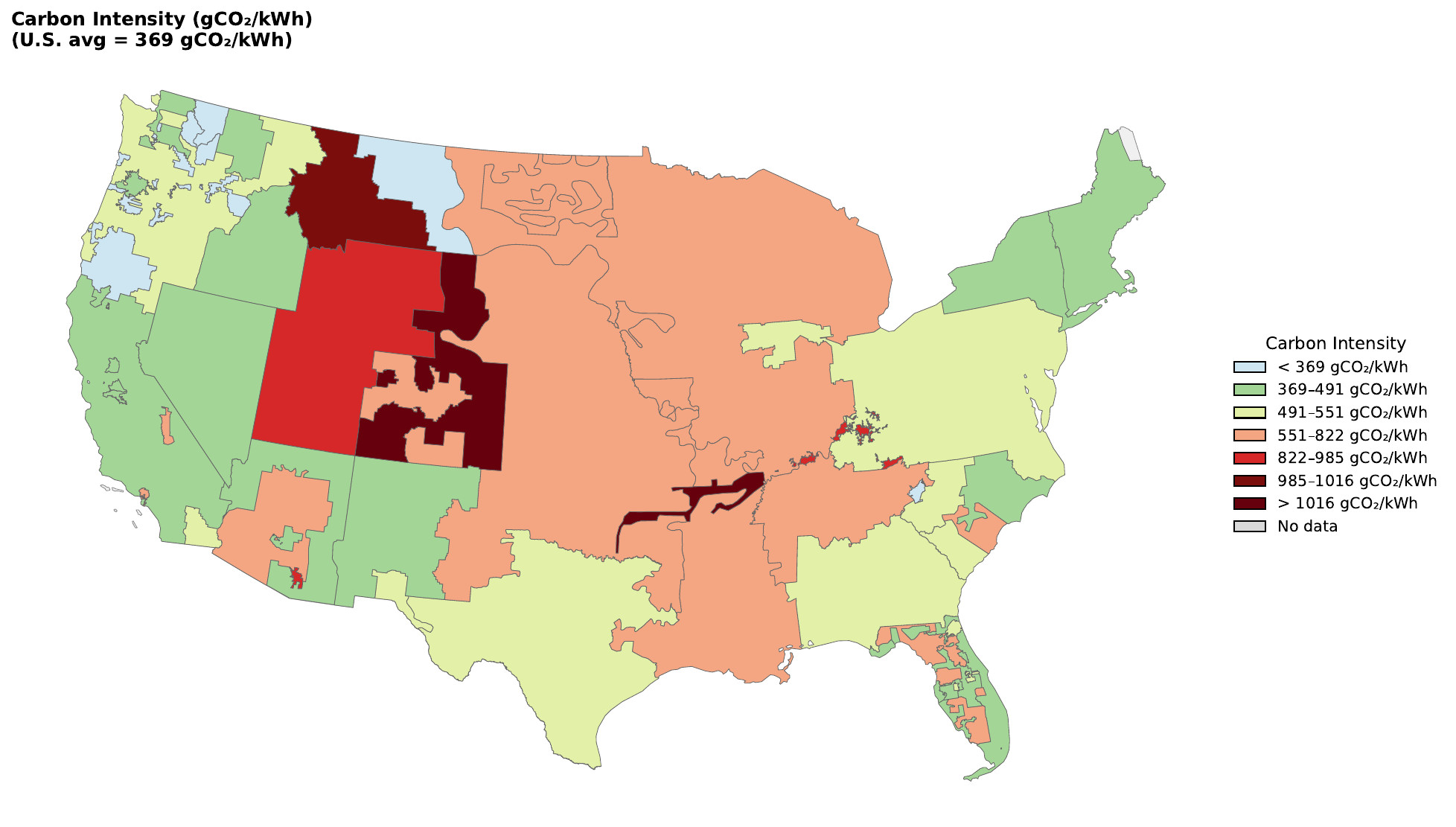}
\end{center}
\caption{\textbf{Carbon intensities of electricity consumption for hyperscale US data centers by balancing authority.} Carbon intensity is defined as the amount of carbon dioxide emissions produced per unit of electricity generated, or consumed, and is expressed in units such as grams of CO$_2$ per kilowatt-hour (gCO$_2$/kWh) for electricity generation. The figure shows HDCs' carbon intensity for electricity consumption at the balancing authority level, in grams of CO$_{2}$ per kWh.} 
\label{fig:carbon_intensities}
\end{figure}

For context, the HDC electricity-weighted carbon intensity is comparable to the national electricity carbon intensities of several relatively carbon-intensive power systems and remains substantially higher than those of lower-carbon European systems such as France, Spain, and the United Kingdom~\cite{ourworldindata_carbon_intensity}(see Fig. \ref{fig:carbon_intensities}).

\begin{figure}
\begin{center}
\includegraphics[width=1.1\textwidth]{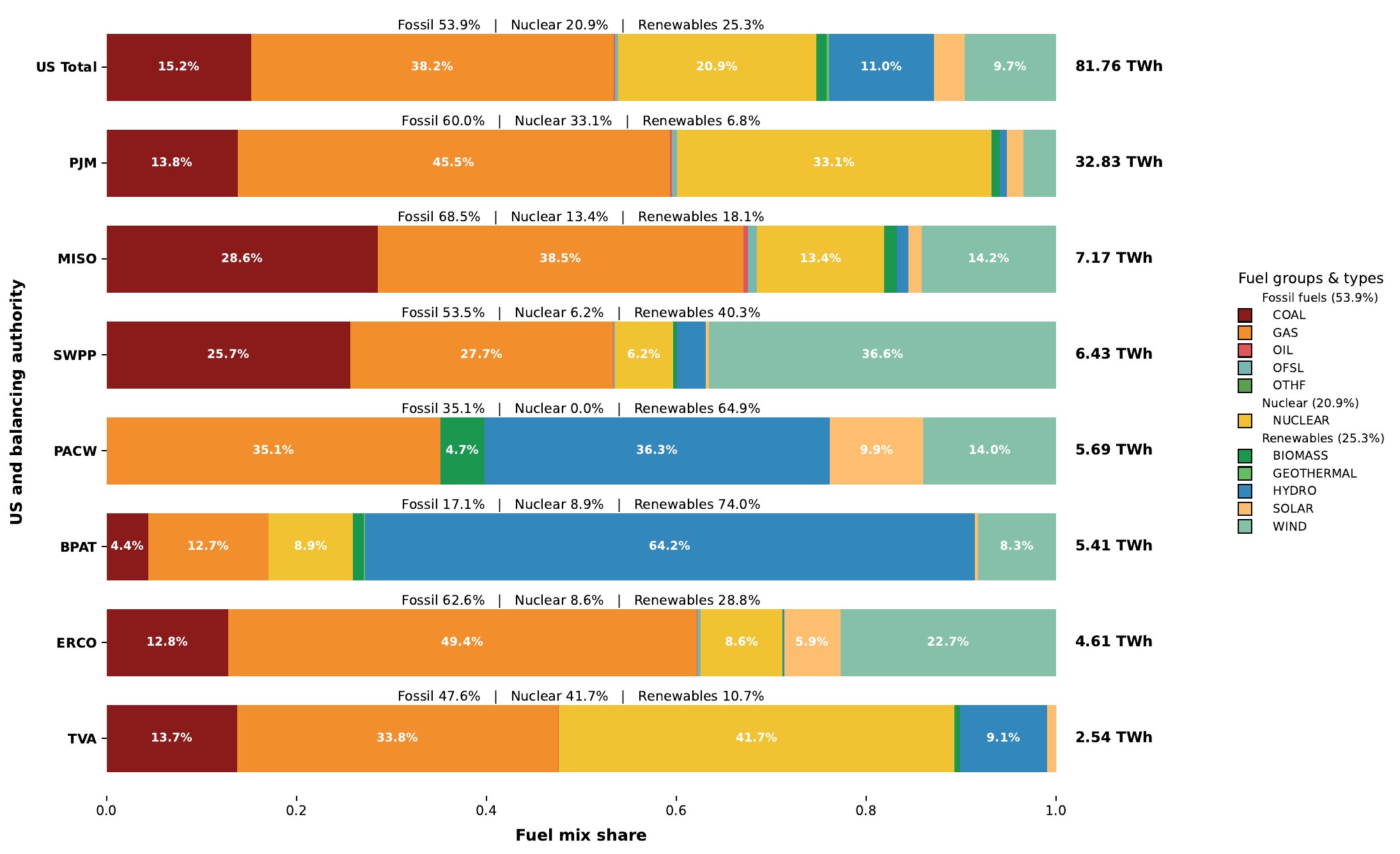 }
%{Images/national_and_top7_regional_fuel_mix_top_us.pdf}
\end{center}
\caption{\textbf{Fuel mix of balancing authorities supplying electricity for US hyperscale data centers.} 
Balancing authorities are ranked by the amount of HDC electricity demand attributed to each region under the central facility-load scenario ($u=0.58$). Bars show the generation-share-weighted fuel mix within each balancing authority, weighted by HDC electricity demand. Under eGRID2023, the national attributed mix is 53.9\% fossil fuels, 20.9\% nuclear, and 25.3\% renewables. For instance, TVA's attributed mix comprises 47.6\% fossil fuels, 41.7\% nuclear, and 10.7\% renewables. We refer to Fig.~S.2.2 for balancing authority regions and corresponding names.}
\label{fig:fuel_mix}
\end{figure}

Carbon intensities are closely related to the fuel mix used in electricity production, which varies significantly depending on the country, region, and types of power plants. The primary components of the fuel mix include fossil fuels (coal, natural gas, and oil), renewable energy (solar, wind, hydro, geothermal, and biomass), and nuclear power. Fig.~\ref{fig:fuel_mix} illustrates the attributed fuel mix of HDC electricity demand nationally and across the largest balancing authorities. Nuclear power provided approximately 20.9\% of attributed HDC electricity, while renewable sources provided approximately 25.3\%. 

PJM is the balancing authority with the largest number of HDCs and the highest HDC electricity demand in our dataset. Under eGRID2023, approximately 60.0\% of PJM's attributed HDC electricity came from fossil-fuel power plants, with coal accounting for 13.8\% and natural gas for 45.5\%. Nuclear represented 33.1\% of the attributed mix, and renewables represented 6.8\%.
 
\subsection*{Publicly Available Web Application}

We developed a data visualization tool that allows users to track the carbon footprint of each HDC included in our analysis. The platform can be found at \texttt{\url{https://www.arcgis.com/apps/dashboards/abc6fbecb1904325bd734392f47a7850}}. The platform offers different geographical layers---balancing authority, state, and county levels---at which the number of HDCs and their electricity consumption, attributable carbon emissions, carbon intensities, and fuel mixes can be explored.

\section*{Discussion}
 
In this paper, we proposed a tool to estimate HDCs' electricity consumption and attributable CO$_2$ emissions. Our data driven approach has the following features; 1) it integrates heterogeneous sources of data on HDCs from various sources (i.e., private data center data providers and web , see Supplementary Materials section \ref{sm:s1}); 2) it validates such data by using satellite images; 3) it allows the assessment of fuel mix thus providing transparent information on carbon intensity; 4) the whole data pipeline is \textit{scalable}, that is, the workflow is fully automated and modular, with each stage implemented as reproducible scripts that operate on tabular/geospatial inputs without manual intervention once the required input columns are provided.

Across facility-load scenarios, we estimate that hyperscale US data centers were associated with approximately 37--54 million metric tons of CO$_2$ over the study period; the central scenario gives approximately 45~Mt. This range is roughly 3.5- to 5-fold the 10.5~Mt peer-reviewed estimate reported for HDCs in 2018 by Siddik et al.~\cite{siddik2021environmental}.

It is important to acknowledge that Shehabi et al.~\cite{Shehabi2024DataCenter} estimate CO$_2$ emissions from all US data centers using server installed-base data from IDC's Worldwide Quarterly Server Tracker for 2003--2023~\cite{idc2023a,idc2023b}. Our approach differs because it focuses specifically on HDCs, uses facility-level power-capacity estimates for 2024--2025, and applies an attributional grid-emissions framework based on EPA eGRID2023 plant-level generation and emissions factors. Under our central scenario ($u=0.58$), we estimate approximately 45~Mt CO$_2$ with a weighted carbon intensity of approximately 0.545~kgCO$_2$/kWh. The higher carbon intensity relative to all-data-center estimates likely reflects the geographic concentration of HDCs in balancing authorities with above-average grid carbon intensity. Our estimate is not directly comparable to Xiao et al.~\cite{xiao2025environmental}, who quantify AI-server-specific emissions rather than the full electricity footprint of hyperscale facilities. Their results therefore provide a lower-scope AI-specific benchmark rather than a like-for-like comparison.

Across our scenarios, approximately 54\% of the electricity attributed to
HDCs is supplied by fossil-fuel generation, 21\% by nuclear, and 25\% by renewables. These shares are computed as the generation-share-weighted fuel mix of supplying plants within each balancing authority, weighted by HDC electricity demand at the BA level (i.e., they reflect attributed electricity-generation share, not plant counts). Within the fossil component, natural gas contributes approximately 38\% of total attributed electricity
and coal approximately 15\%. Among non-fossil sources, nuclear contributes nearly 21\%, while hydro, wind, solar, biomass, and geothermal together account for 25\%.

We also found that approximately 90.1\% of HDC electricity demand is located in balancing authorities with carbon intensity above the US national average, despite the fact that some of the largest HDC operators are also among the largest private purchasers of renewable energy~\cite{bnef_clean_energy}.

Our analyses underscore the importance of relying on data-driven decision-making to inform policies aimed at mitigating the environmental impact of data centers while acknowledging their crucial role as the backbone of modern computing technology. The urgency of this matter is further amplified by the anticipated surge in data center proliferation driven by the rapid advancement and widespread adoption of AI technologies. The AI race significantly increases computational demands and, consequently, electricity consumption of data centers.

Furthermore, our analyses highlight the importance of recognizing data centers and, in particular, HDC facilities, as a relevant sector of the economy. Policy has lagged behind the large growth of data centers, although states and federal regulators are increasingly turning their attention to the industry \cite{datacenter_knowledge_tougher_reporting, uptimeinstitute_mandates_crypto}.

\subsection*{Limitations and Future Research}

Our study has limitations. First, our data do not allow us to differentiate directly between general-purpose and AI-specialized operations at the facility level. As a result, facility-level workload mix, hardware density, and cooling configuration remain partly unobserved. Two important sources of uncertainty act in opposite directions. Incomplete facility coverage biases aggregate electricity and emissions totals downward, because not all identified facilities could be validated and retained in the analytical sample. By contrast, applying a scenario-based facility-load coefficient to total facility electrical capacity may bias totals upward or downward depending on workload mix, design contingency, and the relationship between average operating load and design-case capacity. We therefore interpret our national and state-level estimates as bounded scenario-based estimates rather than as a strict lower bound.  The  IEA's latest report \cite{iea2025energyai} suggests that HDCs account for approximately 60\% of the 200 TWh consumed by all US data centers annually in 2024. Furthermore, AI data centers are the dominant use case for new builds and expansions \cite{Colliers2025DataCenter}.

We do not explicitly differentiate hyperscale facilities by workload type (e.g., AI training, inference, or enterprise computing), as facility-level workload composition is not publicly disclosed. 
These workloads differ in hardware density, cooling configuration, and temporal load profiles, introducing uncertainty at the individual-facility level \cite{chokse2024llmpower}. 
Our analysis therefore focuses on aggregate regional and national electricity consumption and emissions, which are driven primarily by total facility electrical load rather than by facility count. 
Per unit of IT electricity, nearly all consumed energy is ultimately rejected as heat. AI-specialized facilities may therefore have higher thermal flux per unit floor area because of greater power density, but their cooling energy as a share of total facility electricity is not necessarily larger, especially where liquid cooling or rear-door heat exchangers reduce cooling overhead. In the absence of facility-level PUE data, we do not assign separate cooling factors to AI-specialized facilities. Public evidence suggests that such facilities remain relatively few in number, limiting their influence on aggregate results. For AI-specific emissions projections that account for workload characteristics and hardware configurations, we refer readers to \cite{xiao2025environmental}, who employ a supply-chain bottleneck approach to estimate the environmental impacts of AI server deployment.

Another limitation of our analysis is incomplete coverage of US HDCs. We were unable to obtain square footage or power-capacity data for all 675 facilities initially identified. This limitation likely reduces aggregate totals relative to the full U.S.\ hyperscale population, but it does not by itself determine the net direction of bias because facility-load conversion introduces offsetting uncertainty. For this reason, we do not describe the reported totals as a formal lower bound.

%Third, due to a lack of direct measurement data for server power consumption and capacity utilization between data centers, we used average rates of 0.66 for HDCs across the study period and geography. As data center electricity consumption (and therefore emissions levels) vary directly with utilization, a more precise model would factor in a specific uptime for each data center evaluated as part of the study. However, while the non-peer-reviewed LBNL 2024 United States Data Center Energy Usage Report\cite{Shehabi2024DataCenter} had access to direct measurements of facilities, these data are not available.

%Lastly, when estimating power capacities for facilities with missing data, we primarily relied on the square footage of the data centers, as indicated in the literature, as well as their geographic location and climate zone. Factors such as industry economies of scale and regional temperature profiles were considered as potential determinants of power capacities. Although these methods yielded reasonable estimates, achieving a predictive performance of approximately 0.80, the overall precision could be further improved by incorporating additional facility-specific information. However, given that only 6 out of 403 facilities' power capacities are predicted, the impact of this limitation on the overall analysis remains relatively minor.

Overall, there is a broader challenge in the field: the absence of comprehensive, publicly available datasets on HDCs. Unlike other sectors with more transparent reporting standards, data center operators typically do not disclose detailed infrastructure data, and existing studies often rely on proprietary or restricted datasets that are not made available to the research community. 

In this paper, we adopted an \textit{attributional methods} (i.e., a generation-weighted average emission model), to assign each data center the power plant that supplied the electricity demanded \cite{ekvall2005normative,nordenstam2021attributional}. 
Our attributional framework relies on EPA eGRID plant-level generation and emissions data, which are based on plants and generators that provide power to the electric grid and report energy and environmental data to the U.S. government. As a result, behind-the-meter and non-reporting distributed generation may not be fully represented in the plant-level attribution layer. This limitation may lead to local overestimation of carbon intensity in balancing authorities with high levels of distributed renewable generation. At the national scale, however, the magnitude of this omission is bounded by the relatively small share of total U.S. generation from small-scale distributed resources. Because plant-specific emissions factors, fuel inputs, and generation quantities are not reported for smaller units, it is not feasible to incorporate sub-25~MW generators into our analysis. Although this threshold excludes many small renewable installations, national EIA statistics show that the aggregate contribution of these units is modest: small-scale solar ($<1$~MW) accounted for approximately 84,630~GWh of output in 2024 (about 2\% of U.S. electricity generation), with the vast majority of renewable output produced by utility-scale facilities.\cite{eia_electric_power_monthly_2024_smallscale,eia_utility_smallscale_solar_2023} 
Therefore, while this limitation may lead to local overestimation of carbon intensity in balancing authorities with high levels of distributed renewables, its impact on our national emissions estimates is structurally bounded by the reporting constraints of the underlying EPA dataset.\\
Separately, we report sensitivity to grid vintage in Supplementary
Table~\ref{tab:egrid_year_sensitivity}: using eGRID2022 instead of eGRID2023
would shift attributed emissions by approximately +3\% (Mt) and +3\%
(g/kWh) under the central scenario. This shift reflects differences in grid vintage, including updated plant-level generation, emissions rates, and resource-mix inputs in eGRID2023 relative to eGRID2022;
both vintages are documented in the supplement so that readers can
reconstruct either result.

Other approaches, called \textit{consequential},  have been used in the literature on emission accounting \cite{xing2023carbonrespondercoordinatingdemand,jagannadharao2023timeshifting,dodge2022measuringcarbonintensityai, guidi2025environmental}, and serve a different purpose, that is, identify power plants that respond to the \textit{increase} in electricity demand from the establishment of new electricity-hungry facilities (e.g., a new data center) \cite{Gagnon_and_Cole2022}.  These approaches measure the change in total emissions from a change in electricity demand \cite{brander_most_2022} (i.e., the marginal emissions rate).  The responding, or marginal, power plants in these situations generally do not have the same emissions rates as the average of all existing power plants (consequential approach).   When making decisions on when or where to operate a new data center, it will be important to use consequential methods that model the change in total carbon emissions as a consequence of a policy. A more comprehensive approach is left for future research.

A further limitation is that the facility inventory covers May~2024 to April~2025, whereas the plant-level generation and emissions factors are drawn from EPA eGRID2023, the most recent complete plant-level eGRID release available at the time of revision. This one-year lag is necessary because complete plant-level eGRID data are released retrospectively. As a result, our attribution reflects the most recent complete observed grid mix rather than contemporaneous hourly or monthly generation during the exact facility-observation window.

Finally, because our accounting is based on annual average electricity
demand and annual average emissions factors, it does not capture the
temporal correlation between cooling-driven load variation and hourly grid
carbon intensity. Even if IT load were approximately flat, cooling load would vary with ambient temperature and humidity, potentially shifting facility electricity demand toward hours with different grid carbon intensity. Depending on the balancing authority, this could increase or decrease emissions relative to annual-average accounting. Quantifying the direction and magnitude of this interaction requires hourly facility-load and grid-emissions modeling, which is outside the scope of this study.

\section*{Materials and Methods}

Our analysis and corresponding data pipeline can be described in five steps.

First, we compiled a dataset of 675 HDCs across the US, acquiring facility-level data from a data center's data provider (details are provided in the Supplementary Materials). The initial dataset included provider names, geocoded locations, and facility types. In the resulting dataset, we had available information on power capacity for 397 HDCs and square footage information for 392. We validated this dataset through manual verification and satellite imagery analysis using \textit{OpenStreetMap}, ultimately obtaining complete square footage data for 403 facilities. Details, including steps to validate the dataset through OpenStreetMap, are provided in the Supplementary Materials Section \ref{sm:s1}. Figure \ref{fig:location}, shows the geographical location of 403 HDCs in the contiguous US included in our analysis. %Figure \ref{fig:size} shows the distribution of their size in terms of square footage.  

Second, for the 6 out of 403 HDCs with missing power capacity information, we trained a gradient boosted regression tree (GBRT) model \cite{prettenhofer2014gradient} to impute the missing values. The model used building square footage, climate type, and balancing authority region as predictors. 
Performance was evaluated under both a random 85/15 hold-out split and grouped validation schemes that hold out entire balancing authorities, states, or climate categories. Because facilities are geographically clustered, grouped validation is more informative about out-of-region generalization than random splitting; the relative ordering of performance metrics across schemes depends on the structure of held-out variance, as discussed in Supplementary Section~\ref{sm:s2}.

Third, for each HDC, we estimated annual electricity consumption by multiplying total facility electrical capacity by annual hours of operation and by a facility-load coefficient that translates nameplate capacity into average operational electricity draw. Because the appropriate coefficient depends on workload mix, design contingency, and headroom — none of which is observable in our dataset — we report results under four physically motivated scenarios (Section~\ref{sm:s3_utilization}). This is a distinct facility-level approximation required by our data structure; it is not the server-level utilization-plus-PUE framework used by Shehabi et al.~\cite{Shehabi2016DataCenter,Shehabi2024DataCenter}. Details are provided in Supplementary Materials Section~\ref{sm:s3}.

Fourth, we assigned each data center to its balancing authority region and allocated its electricity demand across power plants in that region using an energy-generation-weighted attributional model. We use EPA eGRID2023 plant-level generation, emissions, and fuel-mix data, the most recent complete eGRID release available for plant-level attribution at the time of revision. The HDC observation window (May~2024 to April~2025)
overlaps substantially with calendar-year 2024 grid operation; eGRID2023
became available between our initial submission and this revision and provides the closest publicly available retrospective grid baseline. We
report sensitivity to grid vintage in Supplementary Table~\ref{tab:egrid_year_sensitivity}, which shows that using eGRID2022 instead of eGRID2023 inflates attributed
emissions by approximately 3\% (under the central scenario), reflecting updated plant-level generation, CO$_2$ emission rates, resource-mix inputs, and eGRID processing updates between vintages (e.g.,
PJM combustion CI: 589\,$\rightarrow$\,536~gCO$_2$/kWh; MISO: 694\,$\rightarrow$\,643~gCO$_2$/kWh; SWPP: 805\,$\rightarrow$\,738~gCO$_2$/kWh). Details are provided in Section~\ref{sm:s4}.

Fifth, we computed attributable CO$_2$ emissions and carbon intensity metrics to enable standardized comparisons of the environmental impact of HDC electricity consumption across facilities, states, and balancing authorities. Details are provided in Section~\ref{sm:s5}.

A reproducibility summary documenting all data sources, assumptions, and validation metrics is provided in Supplementary Table~\ref{tab:repro_summary_compact}.

%We gathered and validated detailed information for each HDCs, including the full address, exact location of the building, square footage, and power capacity specifications. Given the lack of open source datasets for HDCs, we built our data pipeline by integrating publicly available data, data obtained by Web scraping, and proprietary data from data center service providers \textit{Baxtel} and \textit{datacenters.com}. We validated the dataset through satellite imagery using \textit{Open Street Map}. Additional details on data collection and validation are provided in the Supplementary Materials.
 
%Power capacity refers to the maximum amount of electricity that equipment in a given data center can draw, expressed in megawatts (MW). This number includes electricity used for computation and networking (\textit{critical IT load}) as well as electricity used for cooling, backup power, and electricity lost due to system inefficiency. For the five data centers with missing power capacity data, we deployed a gradient-boosted regression tree model to estimate their power capacities. We trained the model using features from the 397 data centers with complete data, and then estimated the power capacity of the remaining data centers.  The input of the model included data center features such as square footage, location, balancing authority, and regional climate. The model for imputation of missing data had an R-squared of 0.81 (see Supplementary Materials for details).

\bibliographystyle{Science}
\bibliography{scibib}

\noindent \textbf{Acknowledgments:} We benefited from helpful comments and suggestions from Michelle Audirac, Nat Steinsultz, and Henry Richardson. We wish to thank them for their support and insights.\\

\noindent \textbf{Funding:} National Institute of Environmental Health Sciences-Harvard Center Pilot Project P30ES000002 (FJBS). R01AG066793, R01ES034373. \vspace{1cm}\\
\noindent \textbf{Author contributions}
Conceptualization: FJBS, FD, GG; 
Methodology: FJBS, FD, GG, SD, CS; 
Data Collection and Data Wrangling: GG, KB, JG;
Investigation: FJBS, FD, GG, JG, CS; 
Visualization: FJBS, FD, GG, KB; 
Funding acquisition: FJBS, FD; 
Project administration: FJBS, FD; 
Supervision: FJBS, FD;
Writing – original draft: FJBS, FD, GG, CS;
Writing – review \& editing:  all authors.
\vspace{1cm}\\
\textbf{Competing interests:} The authors declare that they have no competing interests.
\vspace{1cm}\\
\textbf{Data and materials availability:} The dataset we constructed for our analysis will be made available for collaboration with the authors of this work. Requests can be addressed to the last author. 

All code used to preprocess the data, estimate hyperscale data center power capacity, compute electricity use and emissions, and generate all figures is publicly available in the repository \texttt{https://github.com/gianguidi/hyperscale-emissions}. The repository contains (i) a complete and fully specified \texttt{scikit-learn} pipeline for the power-capacity model, including all predictors, preprocessing steps, hyperparameters, and evaluation metrics; (ii) split artifacts documenting the fixed random and grouped validation schemes, including \texttt{splits.json}, which records the fixed random 85/15 split with seed 42 and grouped hold-out definitions by balancing authority, state, and climate category; (iii) a reproducibility note, \texttt{REPRO.md}, documenting the validation sample, imputation rule, random seed, and software versions; (iv) \texttt{reproduce\_model.py}, a lightweight script that reads the split artifact and reports the validation metadata; (v) the full emissions-attribution workflow using EPA eGRID2023; and (vi) figure-generation scripts that reproduce all visualizations in the main text and Supplementary Information.

Because the underlying facility‐level dataset is subject to a data use agreement (DUA), we do not release raw facility identifiers or coordinates. Instead, we provide synthetic examples, anonymized predictors, and aggregated regional datasets that allow reviewers and researchers with similar data to execute the entire pipeline end‐to‐end.

\textbf{Data availability}
The facility‐level dataset used in this study is subject to a data use agreement (DUA) and cannot be publicly released because it contains commercially sensitive information about hyperscale data‐center locations and infrastructure. To support reproducibility, we provide in our public repository (i) synthetic facility‐level datasets that preserve the marginal distributions of key variables while removing all identifying information, 
(ii) anonymized predictors for the capacity‐estimation model, and (iii) aggregated 
state- and balancing authority-level electricity consumption and emissions datasets. These aggregated products contain no facility identifiers or coordinates and satisfy the constraints of the DUA while enabling full execution of the modeling pipeline by external researchers. Publicly available datasets from EPA (eGRID) and EIA (Form 860, Electric Power Monthly) used for emissions and generation attribution can be obtained directly from their respective websites.

Our code is hosted as an open-source project at: \\ \texttt{https://github.com/gianguidi/hyperscale-emissions}.

\vspace{1cm}
\noindent \textbf{Supplementary Materials}\\
Materials and Methods\\
%Robustness checks\\
Supplementary Methods\\
Supplementary Figures\\
Supplementary Tables\\
Supplementary References\\
\pagebreak

\appendix

\counterwithin{figure}{section}
\counterwithin{table}{section}
 \pagenumbering{arabic}
    \setcounter{page}{1}

\makeatletter
\def\@seccntformat#1{\@ifundefined{#1@cntformat}%
   {\csname the#1\endcsname\space}%    default
   {\csname #1@cntformat\endcsname}}%  enable individual control
\newcommand\section@cntformat{\thesection \space} % section-level
\makeatother
\renewcommand{\thesection}{S.\arabic{section}}
\counterwithin{equation}{section}
\counterwithin{figure}{section}
\counterwithin{table}{section}
    
\begin{center}
    \Large \ Supplementary Materials to: \\
    ``Assessing the Carbon Emissions and Energy Consumption of U.S. Hyperscale Data Centers''

    \vspace{0.25cm}
    \large Gianluca Guidi, Francesca Dominici, Callaway Sprinkle,\\ Jonathan Gilmour, Kevin Butler, Eric Bell, Scott Delaney,\\ Falco J. Bargagli-Stoffi
\end{center}

\doublespacing
\thispagestyle{empty}
\tableofcontents
\clearpage

\newpage

\pagenumbering{arabic}
\setcounter{page}{1}

\section*{Introduction}

To construct a comprehensive data platform for United States' (US) hyperscale data centers, we implemented a multi-phase data pipeline involving data collection, validation, prediction, and analysis. 

First, we acquired data on hyperscale data centers from \texttt{Baxtel.com}, a leading information resource for data centers in the US. This dataset comprised 675 hyperscale data centers (HDCs) across the US and included provider names and facility locations. Building square footage and power capacity were available only for a subset of facilities. Data validation was performed through manual review using a multi-source composite satellite and aerial imagery product together with building-footprint data from OpenStreetMap (OSM), enabling us to obtain or validate building-area measurements for the analytical sample. Details are provided in Section~\ref{sm:s1}.

Second, we estimated missing facility power capacities using a Gradient Boosting Regression Tree (GBRT) model. The model used building square footage, climate type, and balancing authority as predictors. Details are provided in Section~\ref{sm:s2}.

Third, for each HDC, we estimated annual electricity consumption by multiplying total facility electrical capacity by annual hours of operation and by a scenario-based facility-load coefficient. This coefficient translates reported facility-level electrical capacity into annual average electricity demand. This is a distinct facility-level approximation required by our data structure; it is not the server-level utilization-plus-PUE framework used by Shehabi et al.~\cite{Shehabi2016DataCenter,Shehabi2024DataCenter}. Because our dataset does not separately observe IT load, cooling load, PUE, or design-contingency margins, we report electricity consumption under four physically motivated facility-load scenarios. Details are provided in Section~\ref{sm:s3}.

Fourth, we assigned each data center to its balancing authority region and allocated its electricity demand across power plants in that region using an energy-generation-weighted attributional model. We use EPA eGRID2023 plant-level generation, emissions, and fuel-mix data, the most recent complete official EPA eGRID plant-level release available at the time of revision. Details are provided in Section~\ref{sm:s4}.

Fifth, we computed attributable CO$_2$ emissions and carbon intensity metrics to enable standardized comparisons of the environmental impact of HDC electricity consumption across facilities, states, and balancing authorities. Details are provided in Section~\ref{sm:s5}.

This multi-phase data pipeline facilitated the creation of a robust and comprehensive data platform for analyzing data centers in the United States, their energy consumption, attributable carbon emissions, and environmental impact. By integrating diverse data sources and employing advanced machine learning techniques, we were able to fill gaps in our dataset and conduct a detailed analysis of US data center power capacities and their environmental impact, while providing a valuable tool for stakeholders aiming to optimize data center operations and minimize their carbon footprint. 

Each of these steps is further detailed in the Sections below.

Throughout the analysis, aggregate electricity consumption and emissions are driven by total facility load rather than by the absolute number of data centers. Because larger hyperscale facilities account for a disproportionate share of total electricity demand, incomplete coverage of smaller facilities has limited influence on regional and national estimates. Accordingly, results are interpreted in terms of aggregate load and emissions rather than facility counts.

\section{Data Collection}\label{sm:s1}

\subsection{Baxtel Data}

The first part of the pipeline involved compiling a relevant dataset of data centers in the US. This dataset was created using multiple data sources and methodologies. Initially, we sought a reliable and comprehensive list of data centers in the US. Due to the absence of an open-source dataset providing this information, we acquired data from Baxtel, a leading information resource for data centers. Baxtel offers comprehensive insights into data centers and provides tools to support informed data center deployments. Baxtel supplied a dataset containing information on 675  hypescale data centers across the US. This dataset included the names of HDC providers, addresses, latitudes, and longitudes. Power capacity data was available for 397 of these data centers, while square footage information was provided for 392 centers. 

All Baxtel's information is independently verified by Baxtel's expert researchers prior to inclusion in its database. The data is gathered through a diverse range of sources, including news outlets, media publications, RSS feeds, satellite imagery, direct communication with providers, and facility audits from some of the largest global interconnection providers. As detailed below, we have further manually validated all the HDCs in our sample.

\subsection{Open Street Map}\label{sm:osm}

Specifically, standardized facility designs and the arrangement of multiple facilities on a campus enable the use of satellite imagery to confirm facilities' location and identity as HDCs. To further validate the obtained and collected HDCs' facilities square footages, and enrich the dataset where information was still missing, we used two data sources. First, a composite satellite and aerial imagery product (\texttt{Esri World Imagery}) with a 3-cm high-definition resolution for the US. This high-resolution imagery enabled detailed examination of building characteristics. Second, polygon building footprint data from \texttt{Open Street Map} (2024-03-29 release). The methodology for determining the sizes of HDC buildings involved several steps. 
    
We began by geocoding the address in the \texttt{Baxtel.com} dataset using a method that places the geocoded point directly on the rooftop (or parcel centroid) of the building associated with an address, rather than on the street or curbside. Next, we extracted all building footprints from \texttt{OpenStreetMap}, which includes over 64 million buildings in the US. 
    
For each building, we calculated its geodesic square footage to reduce any impacts of area distortion due to map projection. Next, we spatially intersected the geocoded address points of the HDCs in our dataset with the \texttt{OpenStreetMap} buildings, selecting those where the rooftop address fell within the building’s footprint. The HDC was assigned the square footage of the building footprint containing it. 
    
\begin{figure}[p]
\centering
\includegraphics[
    width=0.96\textwidth,
    height=0.88\textheight,
    keepaspectratio
]{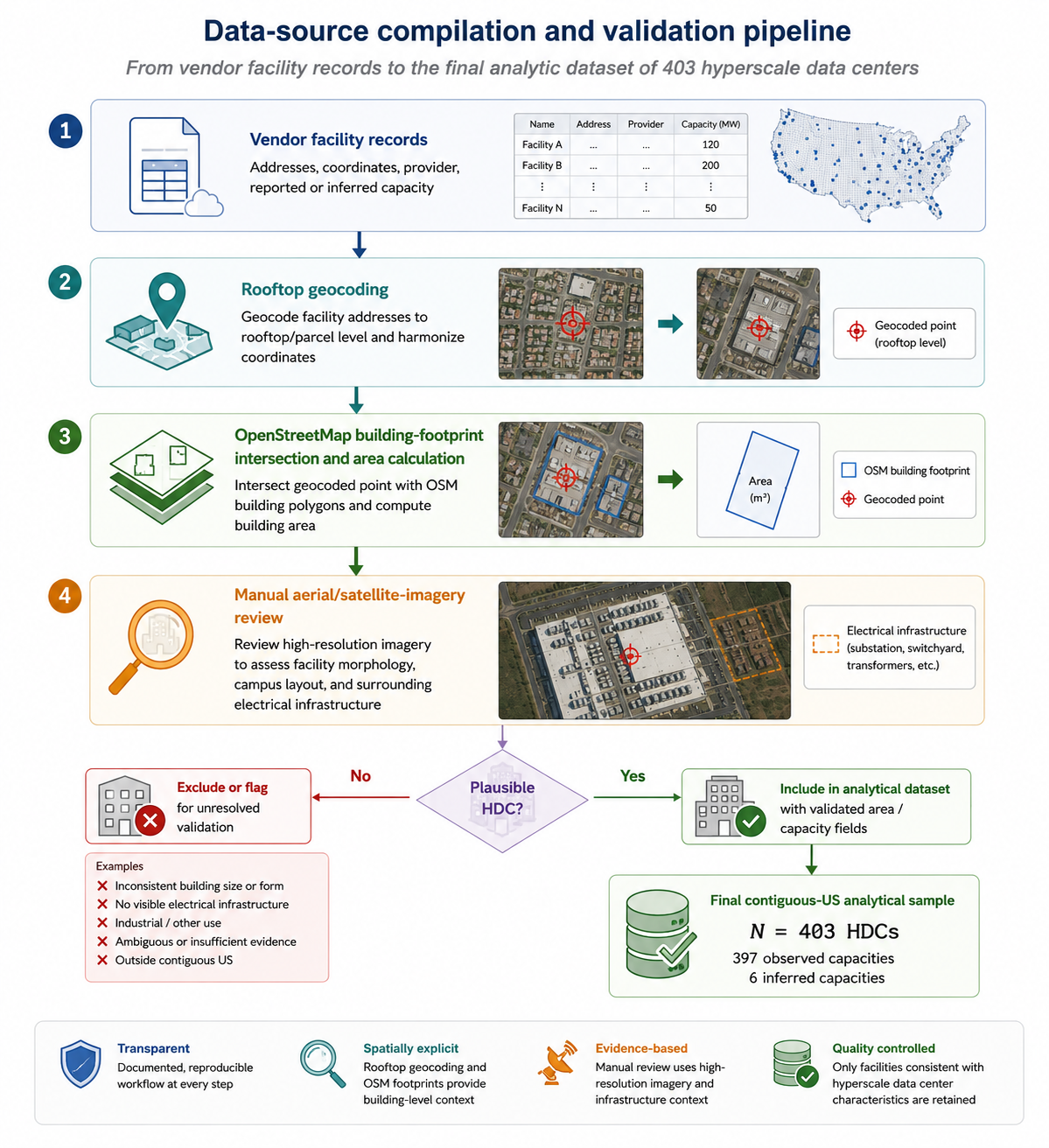}
\caption{\textbf{Data-source compilation and validation pipeline.}
Workflow used to construct the final analytical dataset of 403 hyperscale data centers in the contiguous United States. Vendor-provided facility records were first harmonized and geocoded to rooftop or parcel-level coordinates. Geocoded points were then intersected with OpenStreetMap building footprints to identify candidate data center buildings and calculate building area. Candidate facilities were manually reviewed using high-resolution aerial and satellite imagery to assess building morphology, campus layout, and nearby electrical infrastructure. Facilities were retained when the geocoded point, building footprint, imagery, and infrastructure context were consistent with hyperscale data center use; otherwise, they were excluded or flagged for unresolved validation.}
\label{fig:data_validation_pipeline}
\end{figure}

To validate the geocoding, we examined the high-resolution composite satellite and aerial imagery for the geocoded point and its surrounding geographical area to ensure that the point fell within a building that could reasonably be an HDC. We looked for typical characteristics of HDC facilities such as large, uniform building footprints, multiple similarly sized structures, and clearly visible electrical infrastructure. This process allowed us to determine the square footage for the HDCs for which we had to predict power capacities. 
Figure~\ref{fig:data_validation_pipeline} provides an illustrative overview of this validation workflow, including the sequence from vendor records to geocoding, OSM footprint matching, manual imagery review, inclusion/exclusion decisions, and construction of the final analytical dataset.
    
In total, we were able to obtain square footage information for 403 of the 675 HDCs. Potential reasons why not all facilities could be geolocated and square footage measured include the fact that rooftop geocoding depends on the availability of building or parcel points, which do not exist uniformly across the U.S. and may be outdated; aerial imagery used for validation can also be out of date; the provided address may correspond to a campus building other than the main HDC structure; and given that data centers are being constructed at a rapid pace, some facilities may not yet appear in available imagery or mapping datasets.
    
When we had multiple estimates for the same HDC, we prioritized Baxtel's data over the other estimates, as it was deemed the most accurate due to its more sophisticated and up-to-date retrieval methods. Knowing the exact size of the facilities is crucial to calculating the power capacity estimate, as this information is not directly available.

%\begin{figure}
%\begin{center}
%\includegraphics[width=0.75\textwidth]{Images/infographic.png}
%\end{center}
%\caption{\textbf{The chain of digital services, data centers, and CO$_{2}$ emissions.} This infographic illustrates the interconnected relationship between digital services, data centers, energy consumption, and the consequent emissions from the power plants that supply data centers. It highlights how data centers power digital services by relying on energy from power plants, leading to CO$_2$ emissions that are subsequently released into the environment.}
%\label{fig:diagram}
%\end{figure}

\section{Estimating Power Capacities}\label{sm:s2}

In cases where power capacity data was missing, we estimated it using a machine learning approach. Specifically, we employed a gradient boosting regression tree (GBRT) model, leveraging a variety of features related to data center characteristics and their environments.

Our preliminary analysis, in line with existing literature, confirmed that the power capacity of a data center is highly correlated with its square footage \cite{istrate2024environmental, dante2024,sarkar2024carbon,masanet2020recalibrating,Shehabi2016DataCenter,luers2024will,siddik2021environmental,masanet2020much}. After excluding outliers, defined as data points with a Z-score greater than 2.0, we estimate an average power density of approximately 1,625 watts per square meter (151 watts per square foot). 
This value lies within the broad range reported for modern hyperscale facilities in peer-reviewed and industry sources. We treat this comparison as a plausibility check on the scale of the estimated capacities rather than as an independent validation benchmark for the full emissions workflow. For example, projections from a Goldman Sachs report suggest that data center power densities, driven by the increasing computational demands of AI, may rise significantly, from 1,743 watts per square meter (equal to 162 watts per square foot) today to 1,895 watts per square meter (176 watts per square foot) by 2027, \cite{goldman2023ai}.

\begin{figure}[H]
\begin{center}
\includegraphics[width=0.7\textwidth]{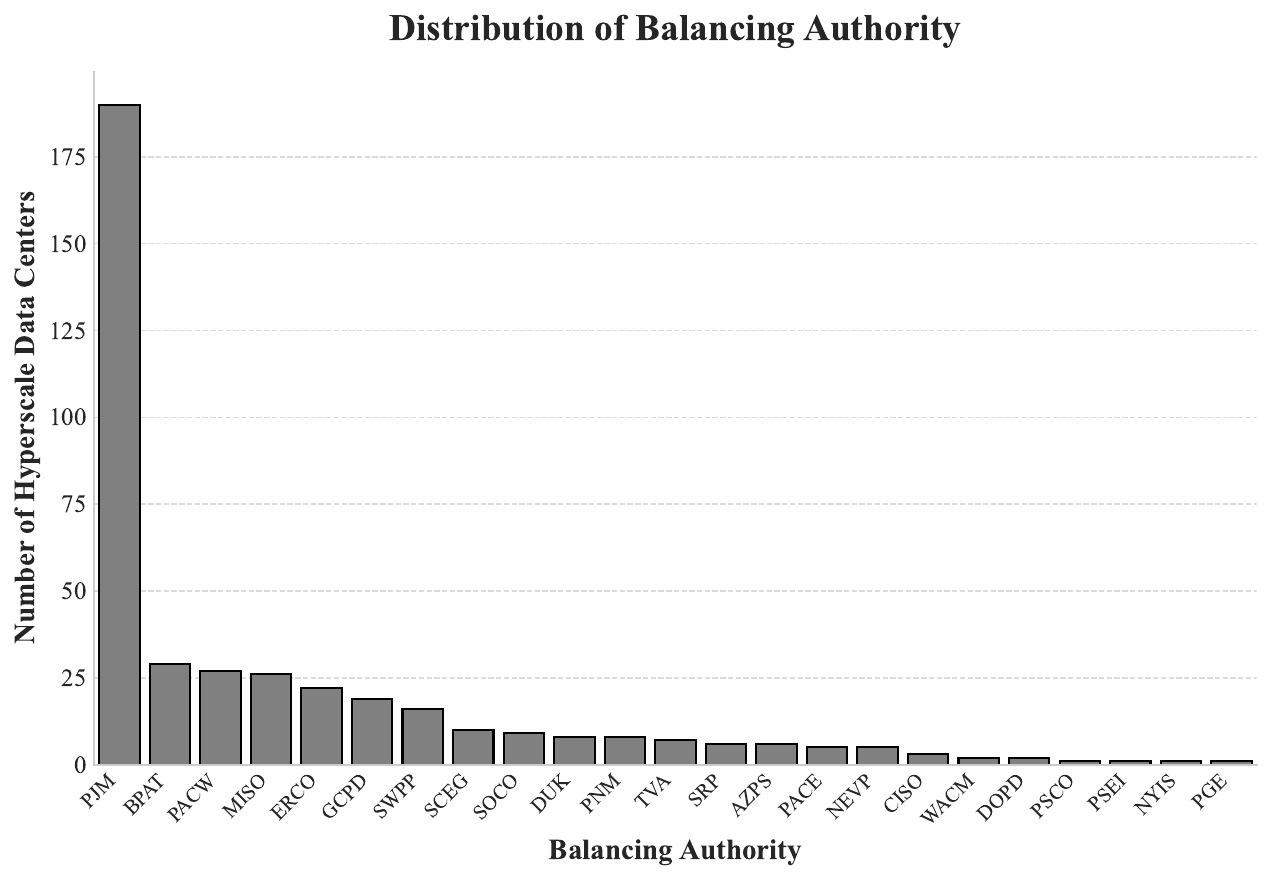}\
\includegraphics[width=0.7\textwidth]{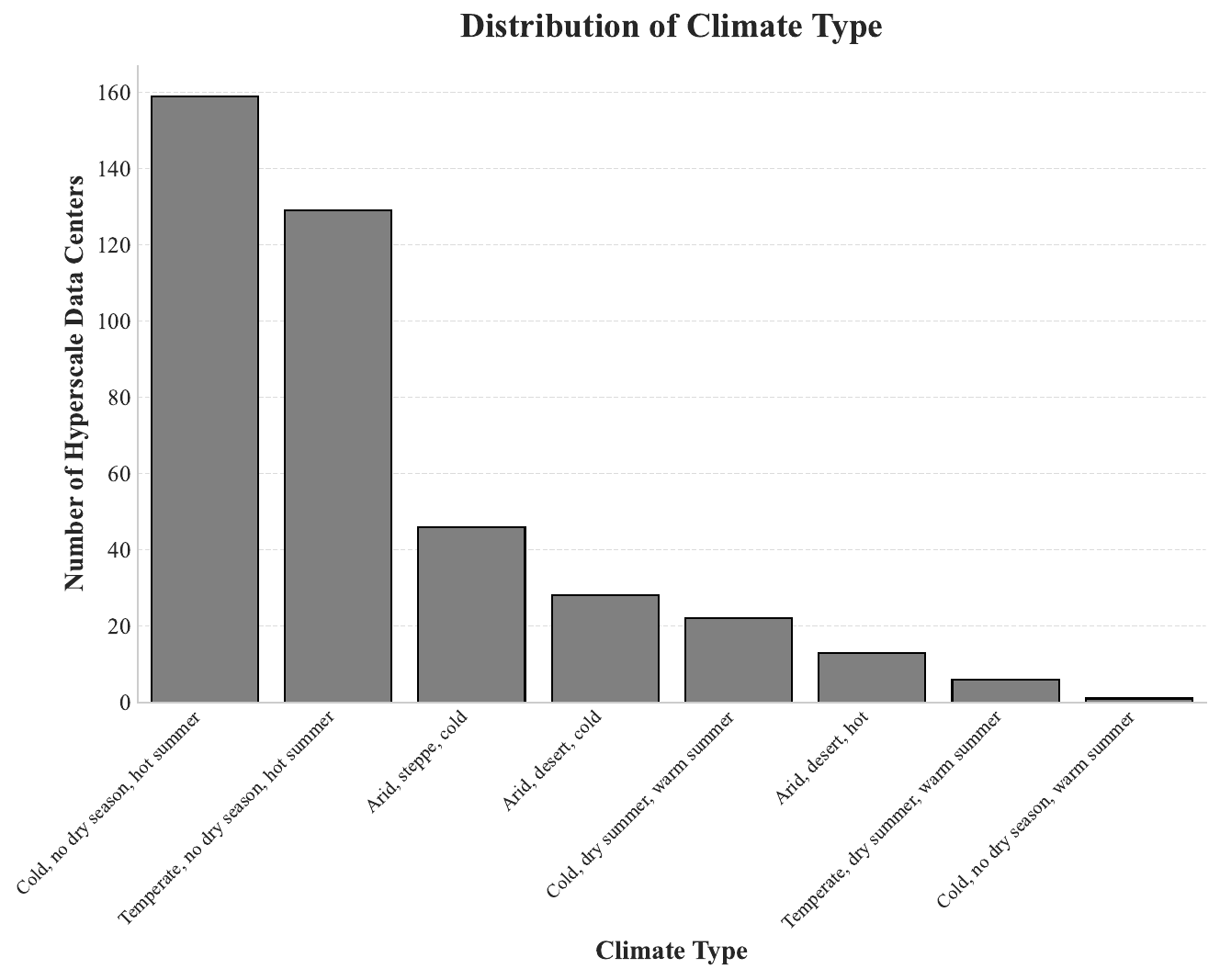}
\end{center}
\caption{\textbf{Data Center Features Categorical Distributions.} 
Top panel: Distribution of hyperscale data centers across balancing authorities. 
PJM contains the largest share of facilities, followed by BPAT and PACW. 
Bottom panel: Distribution of data centers by Köppen-Geiger climate classification. 
Cold climates with no dry season and hot summers are most prevalent, followed by temperate climates with similar characteristics. Arid climate types (steppe and desert) account for a smaller but notable share of facilities.}
\label{fig:features}
\end{figure}

\begin{figure}[H]
\begin{center}
\includegraphics[width=1\textwidth]{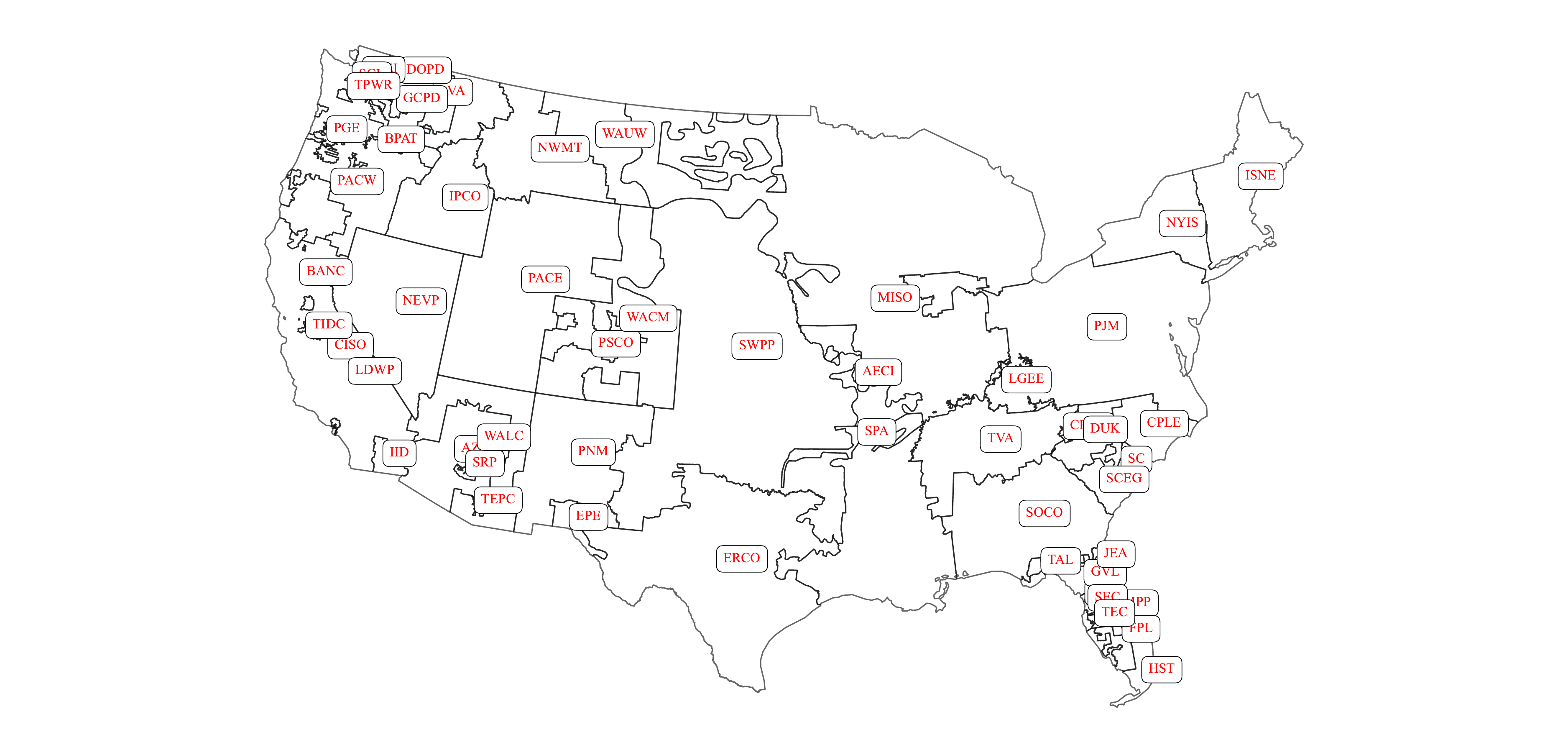}
\end{center}
\caption{\textbf{Balancing Authorities Regions in the US.} Map displaying the geographic boundaries of Balancing Authorities across the contiguous United States. Each region is labeled with its corresponding abbreviation.}
\label{fig:BA}
\end{figure}

The features used in our model included:
\begin{itemize}

\item Climate Type: The climate category of the region where the data center is located, as different climates can affect cooling requirements and, consequently, power consumption. This information was retrieved from Nasa Earth data, at the link \url{https://webmap.ornl.gov/ogcdown/dataset.jsp?dg_id=10012_1$}. Distribution of hyperscale data centers by climate type is depicted in the bottom panel of Figure~\ref{fig:features}.

\item Balancing Authority: The entity responsible for ensuring a balance between electricity supply and demand in the region, which can influence the operational characteristics of data centers. Distribution of hyperscale data centers by balancing authority is depicted in the top panel of Figure~\ref{fig:features}. Balancing authority regions in the US are mapped in Figure~\ref{fig:BA}.

%\item Data Center Type: A categorical variable indicating whether the data center is a hyperscale facility or not. HDCs often have different design and operational standards compared to smaller facilities. For the model, all non-HDCs were grouped together. 

\item Square Footage: The total building square footage of the data center, a critical factor in determining power requirements.

\end{itemize}

To predict the missing power capacities, we split the data that included all known power capacities into a training set and a testing set. 

We used the following preprocessing steps: Scaling and Encoding. Numerical features were standardized to have a mean of zero and a standard deviation of one. Categorical features were converted into one-hot encoded vectors to allow the model to interpret them correctly.
The GBRT model was fit within a fully specified preprocessing pipeline. We report one fixed random 85/15 train--test split (seed 42) only as an illustrative baseline and complement it with grouped validation schemes that hold out entire balancing authorities, states, or climate categories. Because facilities are geographically clustered, grouped validation is more informative about out-of-region generalization than a single random split. We therefore present validation results neutrally and interpret them as sensitivity diagnostics rather than as proof of model performance.
For the illustrative fixed 85/15 split, the model yields $R^2 = 0.807$, MAE = 8.13~MW, and test RMSE = 12.49~MW. Repeated-random and grouped validation results are reported separately in Table~\ref{tab:grouped_validation}. The grouped schemes show that predictive performance depends on the split structure, especially when entire climate categories are held out. We therefore treat the capacity model as a practical imputation tool for a small number of missing facilities rather than as a general-purpose predictive model for out-of-sample hyperscale siting.

%\begin{figure}
%\begin{center}
%\includegraphics[width=0.6\textwidth]{Images/hyperscalers_sqm_distributions.pdf}
%\end{center}
%\caption{\textbf{Hyperscale Data Centers Square Meters Distributions.}}
%\label{fig:size}
%\end{figure}

\begin{figure}[H]
\begin{center}
\includegraphics[width=0.7\textwidth]{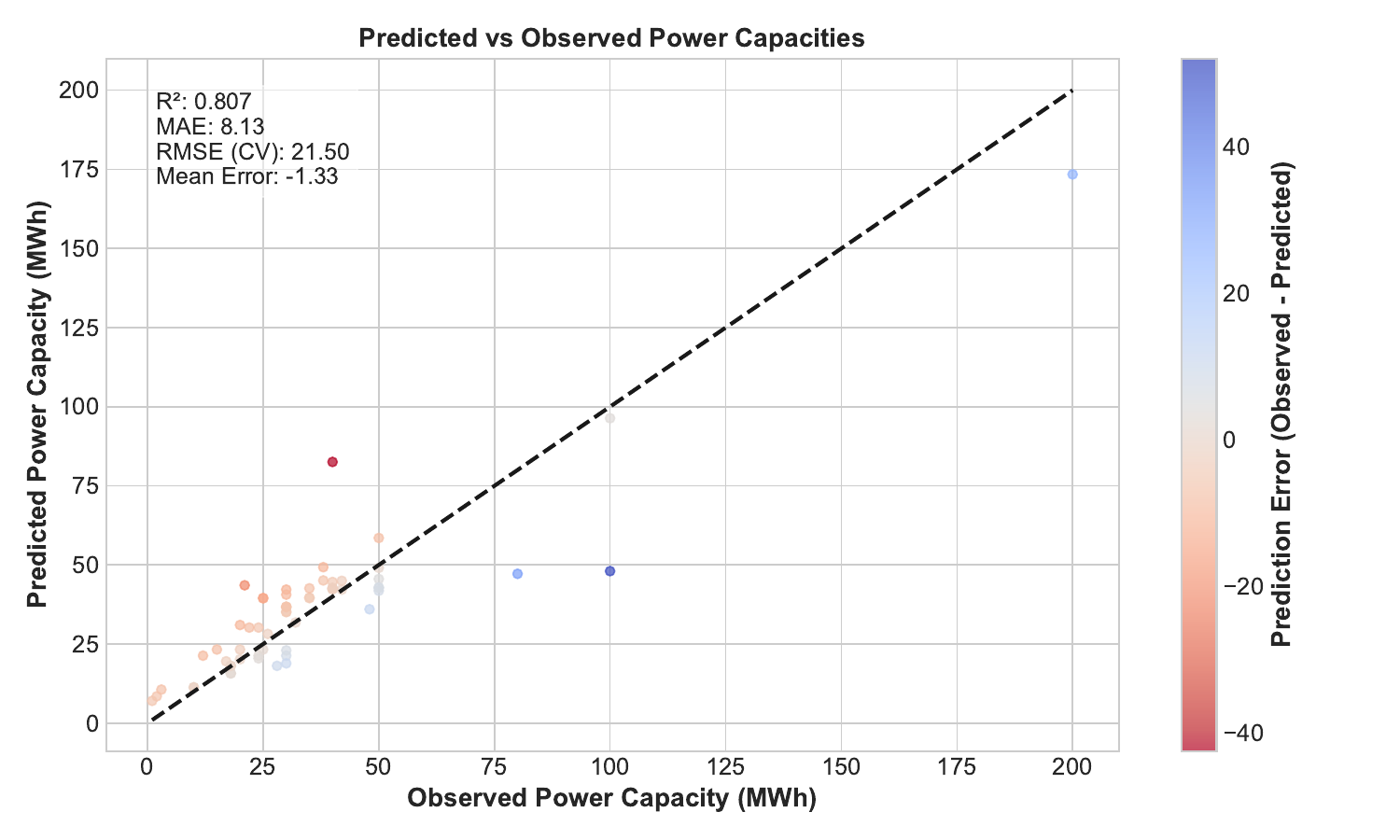}
\end{center}
\caption{\textbf{Data Center Power Capacity Estimation Model Performance.}
Scatter plot comparing predicted versus observed power capacity (MW) for one illustrative random 85/15 train--test split (seed 42). The dashed line indicates perfect prediction. Points are colored by prediction error (observed minus predicted), with blue indicating underprediction and red indicating overprediction. Illustrative split metrics: $R^2 = 0.807$, MAE = 8.13 MW, test RMSE = 12.49 MW, mean bias = $-1.33$ MW. Repeated-random and grouped validation results are reported separately in Table~\ref{tab:grouped_validation}.}
\label{fig:model_performance_1}
\end{figure}

\begin{table}[H]
\centering
\caption{\textbf{State-level CO$_2$ emissions (Mt) across facility-load
scenarios.} Top 10 states by central-scenario emissions. All values use
EPA eGRID2023 plant-level data.}
\label{tab:state_scenarios}
\begin{tabular}{lcccc}
\toprule
State & $u=0.48$ & $u=0.58$ & $u=0.663$ & $u=0.70$ \\
\midrule
Virginia  &  9.5 & 11.5 & 13.1 & 13.8 \\
Oregon    &  4.1 &  4.9 &  5.6 &  5.9 \\
Ohio      &  3.9 &  4.7 &  5.4 &  5.7 \\
Iowa      &  3.8 &  4.6 &  5.3 &  5.6 \\
Nebraska  &  2.2 &  2.6 &  3.0 &  3.1 \\
Texas     &  2.0 &  2.4 &  2.8 &  3.0 \\
Oklahoma  &  1.8 &  2.1 &  2.4 &  2.6 \\
Arizona   &  1.6 &  1.9 &  2.2 &  2.3 \\
Wyoming   &  1.1 &  1.3 &  1.5 &  1.5 \\
Illinois  &  1.0 &  1.3 &  1.4 &  1.5 \\
\bottomrule
\end{tabular}
\end{table}

\paragraph{Model specification and validation.}

Power capacity data were available for 397 of the 403 hyperscale data centers. The remaining facilities lacked reported power capacity but had building square footage information and were therefore included in the prediction set. Building square footage was missing for a small subset of facilities and was imputed using the mean value within the corresponding balancing authority prior to model fitting.

The Gradient Boosting Regression Tree (GBRT) model was fit within a fully specified preprocessing pipeline. We report one fixed random 85/15 train--test split (seed 42) only as an illustrative baseline (Fig.~\ref{fig:model_performance_1}) and evaluate out-of-region generalization using grouped validation schemes that hold out entire balancing authorities, states, or climate categories (Table~\ref{tab:grouped_validation}). All preprocessing steps were embedded within the training pipeline to prevent information leakage.

The feature set included building square footage, climate category, and balancing authority. Numerical features were standardized to zero mean and unit variance, and categorical features were one-hot encoded. Model performance was assessed using the coefficient of determination ($R^2$), mean absolute error (MAE), root mean squared error (RMSE), signed bias, and the shares of under- and over-prediction.

\paragraph{Climate category aggregation.}
While Fig.~\ref{fig:features} reports the full Köppen–Geiger climate classification for hyperscale data centers, for the purpose of validating power-density estimates we aggregate 
climate types into three broad categories: \emph{hot (all year long)}, \emph{hot summer}, and \emph{cold (all year long)}. This aggregation is intentional and designed to maximize 
contrast between thermally distinct environments, rather than to preserve fine-grained climatological distinctions. By grouping the most extreme climate regimes, we test whether 
power-density estimates exhibit implausible sensitivity to climate-driven cooling demands. More granular climate distinctions are shown in Table~S.2.1.

\paragraph{Plausibility check using power density and climate.}To assess whether the climate-dependent component of the capacity estimation model yields operationally plausible results, we compute facility power density (power capacity per unit floor area) and summarize its distribution by climate category (Table~\ref{tab:power_density_climate}). This approach is consistent with \cite{lei2022climatepue}, who show that PUE can vary across U.S.\ climate zones depending on cooling technology and operating conditions. Across climate zones, median power densities are tightly clustered, ranging from approximately 1.5 to 1.6~kW/m$^{2}$. Between-climate differences are small relative to within-climate variability, as reflected by broad interquartile ranges.
This pattern indicates that the climate variable does not drive implausible variation in estimated facility electricity demand, but instead captures broad differences in facility design and cooling infrastructure in a reduced-form manner. The implied power densities are consistent with reported values for modern hyperscale data centers in the literature, supporting the operational plausibility of the model despite the absence of explicit disaggregation between IT and cooling loads.
We therefore interpret these densities as a broad plausibility check on the order of magnitude of the facility-capacity estimates, not as a stand-alone validation of the capacity model or downstream emissions estimates.

\begin{table}[H]
\centering
\caption{\textbf{Facility power density by climate category.} 
Summary statistics of total facility power density, defined as power capacity divided by building floor area and expressed in MW/m$^{2}$.}
\label{tab:power_density_climate}
\begin{tabular}{lcccc}
\toprule
Climate category & Median (MW/m$^{2}$) & 25th pct. & 75th pct. & $n$ \\
\midrule
Hot (all year long)       & 0.00155 & 0.00119 & 0.00175 &  13 \\
Hot summer & 0.00160 & 0.00116 & 0.00210 & 276 \\
Cold (all year long)      & 0.00154 & 0.00104 & 0.00256 &  98 \\

\bottomrule
\end{tabular}
\end{table}

\begin{sidewaystable}[p]
\scriptsize
\setlength{\tabcolsep}{3pt}
\renewcommand{\arraystretch}{0.92}
\centering
\caption{\textbf{Reproducibility summary: data sources, assumptions, headline totals, and ML validation.}}
\label{tab:repro_summary_compact}

\begin{tabular}{p{3.2cm} p{4.2cm} p{14.2cm}}
\toprule
\textbf{Block} & \textbf{Item} & \textbf{Definition / value (units inline)} \\
\midrule

\textbf{Facility dataset} &
HDC count &
$N=403$ hyperscale data centers after validation \\

&
Facility power capacity& 
Total facility electrical capacity (MW), interpreted as meter/infrastructure capacity including IT + cooling + electrical overhead$^{a}$ \\

&
Building area &
Median $2.04\times 10^{4}$ m$^2$; IQR $[1.39,\,2.93]\times 10^{4}$ m$^2$ \\

&
Balancing authority , climate category &
Categorical identifiers used for attribution (balancing authority) and reduced-form geographic/cooling proxy (climate) \\

\midrule
\textbf{Electricity use} &
Annual hours &
$H=8760$ h/yr \\

&
Scenario coefficients & $u \in \{0.48, 0.58, 0.663, 0.70\}$; central reference $= 0.58^{b}$\\

&
Facility annual electricity &
$E_i(u)=\texttt{current\_mw}_i \times H \times u$ (MWh/yr) \\

National totals & 67.7 / 81.8 / 93.5 / 98.6 TWh across $u = 0.48, 0.58, 0.663, 0.70$\\

&
Coverage vs IEA & Central scenario: $81.8/120 \approx 0.68$ (68\%); range across scenarios: 56--82\%\\

&
Distribution summaries &
Annual electricity (GWh/yr): median 209; IQR [128, 290]. Power density (kW/m$^2$): median 1.56; IQR [1.11, 2.14]. \\

\midrule
\textbf{Emissions attribution} &
Accounting basis &
Location-based: reflects physical generation mix supplying each BA using EPA eGRID2023; PPAs excluded unless independently verifiable 1:1 temporal matching$^{c}$ \\

&
Plant generation + emissions factors &
EPA eGRID2023 Revision 2: plant annual net generation $G(j,B)$, resource mix, and emissions factors $EF(j,B)$ (gCO$_2$/kWh), using year-2023 plant-level data \\

&
Reporting threshold &
eGRID plant-level coefficients available for generators $\ge$25 MW; sub-25 MW units not included due to missing plant-level factors \\

&
Attribution method &
Energy-generation-weighted within each BA: $COEFF_{EGW}(j,B)=G(j,B)/\sum_{j}G(j,B)$; balancing authority carbon intensity is load-weighted mean of plant $EF$ \\

\midrule
\textbf{Capacity model (GBRT)} &
Missingness handled &
Train on facilities with observed \texttt{current\_mw}; impute missing \texttt{current\_mw}. Negative predictions truncated to 0. \\

&
Features &
Numeric: building area / sqft (standardized). Categorical (one-hot): climate, balancing authority \\

&
Split + validation &
Illustrative single 85/15 train--test split (seed 42; Fig.~\ref{fig:model_performance_1}), grouped validation by balancing authority, state, and climate category (Table~\ref{tab:grouped_validation}), and public split artifacts in \texttt{splits.json} \\

&
Performance &
Illustrative single split: $R^2=0.807$, test RMSE = 12.49 MW, MAE = 8.13 MW, bias = $-1.33$ MW. Grouped-validation metrics are reported separately in Table~\ref{tab:grouped_validation}. \\

\bottomrule
\end{tabular}

\vspace{0.15cm}
\raggedright
\scriptsize
\textbf{Notes.}
$^{a}$ \texttt{current\_mw} is treated as total facility electrical capacity (not IT-only).
$^{b}$ We report four scenario coefficients ($u \in \{0.48, 0.58, 0.663, 0.70\}$). The central scenario ($u=0.58$) is informed by the Newkirk et al.\ bottom-up power-flow modeling and our independent Lei--Masanet PUE-implied check. The case $u=0.663$ is retained as a continuity scenario relative to prior bottom-up estimates; the AI-weighted high case ($u=0.70$) brackets AI-training-heavy fleets.
$^{c}$ Contractual instruments excluded without verified time-matching.

\end{sidewaystable}

\paragraph{Error diagnostics by geography and fuel mix}

To assess whether the power-capacity estimation model exhibits systematic bias by geography or generation context, we examined residuals (observed minus predicted capacity) on the held-out test set across balancing authorities (BAs) and dominant regional fuel mix categories.

Across BAs with sufficient test observations ($n \geq 8$), mean prediction bias remains within $\pm$15~MW, and residuals are centered near zero. Larger deviations occur only in BAs represented by very small numbers of facilities, where residual statistics are inherently noisy and not indicative of systematic bias.

We additionally grouped BAs by dominant fuel mix (fossil-dominant, nuclear-dominant, renewables-dominant) using the share of annual net generation by fuel type. Fossil-dominant regions show near-zero mean bias, while renewables-dominant regions exhibit a modest negative bias; however, this group comprises relatively few test observations. Correlation analysis between residuals and regional fossil share indicates a moderate association, but this relationship is driven by small-sample regions and does not materially affect predictions in BAs with substantial facility representation.

These diagnostics indicate that model errors are not systematically structured by geography or fuel mix and instead reflect facility-level heterogeneity, consistent with the aggregate scope of the analysis.

\begin{table}[H]
\centering
\caption{\textbf{Power-capacity model error diagnostics by geography and fuel mix.}
Residuals are computed as observed minus predicted facility power capacity (MW) on the held-out test set.}
\label{tab:error_diagnostics}
\begin{tabular}{lcccc}
\toprule
\multicolumn{5}{l}{\textbf{Panel A: Balancing authority–level residuals (test set)}} \\
\midrule
Balancing authority & $n$ & Mean residual (MW) & Median residual (MW) & MAE (MW) \\
\midrule
PJM  & 25 & $-2.09$ & $-2.99$ & 6.04 \\
BPAT & 5  & $-11.66$ & $-14.50$ & 11.66 \\
MISO & 5  & 3.55 & $-2.39$ & 9.57 \\
GCPD & 5  & 0.45 & $-1.32$ & 5.75 \\
ERCOT & 4 & $-1.91$ & $-1.13$ & 4.16 \\
\midrule
\multicolumn{5}{p{0.9\linewidth}}{\footnotesize
Only balancing authorities with at least four test observations are shown to avoid unstable statistics driven by small samples. No balancing authority with $n \geq 8$ exhibits a mean bias exceeding $\pm$15~MW.}
\\
\midrule
\multicolumn{5}{l}{\textbf{Panel B: Residuals by dominant regional fuel mix}} \\
\midrule
Dominant fuel group & $n$ & Mean residual (MW) & Median residual (MW) & MAE (MW) \\
\midrule
Fossil-dominant      & 37 & $-0.13$ & $-2.39$ & 6.83 \\
Renewables-dominant  & 7  & $-7.55$ & $-6.50$ & 9.88 \\
\midrule
\multicolumn{5}{p{0.9\linewidth}}{\footnotesize
Fuel-group classification is based on the dominant share of annual net generation within each balancing authority. Differences across fuel groups are modest and limited by sample size in renewables-dominant regions.}
\\
\bottomrule
\end{tabular}
\end{table}

Because only six facilities required imputation of power capacity, we also recomputed the main electricity and emissions totals using only the 397 facilities with observed capacity. Headline national and state-level results changed only modestly under this complete-case restriction, indicating that the principal findings are not driven by the imputed subset.

\paragraph{Complete-case sensitivity.}
Power capacity was directly observed for 397 of the 403 facilities; the
remaining 6 facilities had missing capacity at the data-collection stage
and were imputed using the GBRT model described above. To assess whether
the principal findings depend materially on the imputed subset, we
recomputed central-scenario ($u = 0.58$) headline totals using only the
397 facilities with directly observed capacity
(Table~\ref{tab:complete_case_sensitivity}). Total electricity consumption
changes by approximately $-1\%$ and attributable CO$_2$ emissions by
approximately $-1\%$ relative to the full 403-facility sample. The
weighted carbon intensity is essentially unchanged. The principal findings
are therefore not driven by the imputed subset.

\begin{table}[H]
\centering
\caption{\textbf{Complete-case sensitivity of headline national totals.}
Headline totals recomputed under the central scenario ($u=0.58$), with
and without the 6 facilities whose power capacity was imputed.}
\label{tab:complete_case_sensitivity}
\begin{tabular}{lcccc}
\toprule
Sample & $N$ & Electricity (TWh) & CO$_2$ (Mt) & CI (gCO$_2$/kWh) \\
\midrule
All contiguous-US facilities & 403 & 81.8 & 44.6 & 545 \\
Observed capacity only       & 397 & 81.0 & 44.2 & 545 \\
\bottomrule
\end{tabular}
\end{table}

\begin{table}[H]
\centering
\small
\setlength{\tabcolsep}{3.5pt}
\renewcommand{\arraystretch}{1.15}
\caption{\textbf{Capacity-model validation under random and grouped split schemes.}
Performance metrics are reported for a repeated random 85/15 hold-out baseline and for grouped validation schemes that hold out entire balancing authorities, states, or climate categories. Underprediction and overprediction rates are defined as the percentage of held-out facilities for which predicted capacity is below or above observed capacity, respectively.}
\label{tab:grouped_validation}

\begin{tabularx}{\textwidth}{
    >{\raggedright\arraybackslash}p{3.0cm}
    >{\raggedright\arraybackslash}p{3.2cm}
    >{\centering\arraybackslash}p{1.05cm}
    >{\centering\arraybackslash}p{1.15cm}
    >{\centering\arraybackslash}p{1.15cm}
    >{\centering\arraybackslash}p{1.15cm}
    >{\centering\arraybackslash}p{1.45cm}
    >{\centering\arraybackslash}p{1.45cm}
}
\toprule
\textbf{Validation scheme} &
\textbf{Held-out unit} &
\makecell{\textbf{$R^2$}} &
\makecell{\textbf{RMSE}\\\textbf{(MW)}} &
\makecell{\textbf{MAE}\\\textbf{(MW)}} &
\makecell{\textbf{Bias}\\\textbf{(MW)}} &
\makecell{\textbf{Under-}\\\textbf{pred.}\\\textbf{(\%)}} &
\makecell{\textbf{Over-}\\\textbf{pred.}\\\textbf{(\%)}} \\
\midrule
Repeated random 85/15
& Individual facilities; random 15\% hold-out
& 0.491 & 19.95 & 10.40 &  0.27 & 48.2 & 51.8 \\

GroupKFold by BA
& Balancing authority
& 0.667 & 18.00 & 12.33 & -2.12 & 40.3 & 59.7 \\

GroupKFold by state
& State
& 0.609 & 18.38 & 11.83 & -1.08 & 42.2 & 57.8 \\

Leave-one-climate-out
& Climate category
& 0.279 & 27.15 & 13.51 &  1.56 & 48.1 & 51.9 \\
\bottomrule
\end{tabularx}
\end{table}

\paragraph{Interpreting the cross-validation comparison.} The grouped CV
schemes yield higher $R^2$ than the random 85/15 split (0.667 BA-held-out vs
0.491 random; 0.609 state-held-out). This counter-intuitive direction
reflects two competing effects: BA folds reduce within-fold target variance
(facilities within a single BA tend to have similar capacity distributions),
while random splits incur higher variance from leakage of correlated nearby
facilities into both train and test. Climate-held-out CV ($R^2 = 0.279$),
with larger fold sizes and greater out-of-distribution character, is the
most conservative estimate of out-of-region generalization. We interpret
the climate-held-out result as the appropriate benchmark for predictive
uncertainty about new facilities in unobserved geographies and treat the
random-split metric only as an illustrative reference.

Grouped validation indicates that generalization to held-out geographic regions is noisier than suggested by the illustrative single 85/15 split shown in Fig.~\ref{fig:model_performance_1}. Relative to the repeated-random hold-out baseline averaged over seeds, grouped validation is of comparable magnitude for balancing-authority and state-based splits and weaker for leave-one-climate-out validation, reflecting the limited number and broad heterogeneity of climate categories. We therefore treat the fixed random 85/15 split as illustrative and the grouped schemes as the more informative assessment of out-of-region generalization.

\section{Computing Data Center Energy Consumption }\label{sm:s3}

The last part of the pipeline consisted of computing the annual energy load of each data center, distributing this energy load amongst its supplying power plants, computing the CO$_2$ emissions related to those energy load supplies, and calculating the carbon intensity. 

In this study, $\text{Power Capacity}_{i}$ refers to the \emph{total facility electrical capacity} 
(i.e., the maximum deliverable load at the facility meter or electrical infrastructure), 
which implicitly includes both IT (critical) load and cooling/infrastructure overhead. 
As a result, the annual energy load computed in this section represents total facility 
electricity consumption rather than IT-only consumption.

\paragraph{Definition of facility power capacity and utilization.}
Throughout this study, the facility power parameter refers to \emph{total facility electrical capacity}, defined as the maximum deliverable load at the facility’s electrical infrastructure (often referred to as nameplate or meter-level capacity). This quantity implicitly includes both IT (critical) load and cooling and electrical distribution overhead. It does not represent installed IT-only capacity, manufacturer-rated server power, or an operationally derated runtime load.

Utilization is defined as the average fraction of this total facility capacity drawn over the year, reflecting temporal variation in computing activity and idle power draw. Cooling and infrastructure overhead are therefore embedded in the facility capacity term rather than estimated separately. This definition is consistent with the manner in which power capacity is commonly reported in industry-facing datasets and ensures that annual electricity estimates correspond to total facility electricity consumption.
\begin{quote}
\textbf{Terminology clarification.}  
\emph{Total facility electrical capacity}: maximum deliverable electrical load at the facility meter, including IT and cooling/infrastructure.  
\emph{IT load}: electricity delivered to computing equipment only (not modeled separately).  
\emph{Utilization}: average fraction of total facility capacity drawn over the year.
\end{quote}

\subsection{Facility-load Scenarios}
\label{sm:s3_utilization}

The \textit{facility-load coefficient} $u \in [0, 1]$ translates reported
facility power capacity into average annual electricity draw. A value of $u=1$
would indicate continuous operation at nameplate capacity; values closer to $0$
reflect proportionally lower average draw. This coefficient aggregates IT
load, cooling load, and distribution overhead, because our input data report a
single facility-level nameplate capacity rather than disaggregated IT and
cooling loads. Adopting a facility-level coefficient is therefore a pragmatic
choice conditioned on data availability rather than a direct physical
measurement.

Estimates of server utilization and facility operational uptime vary widely
across sources and years. Academic and industry metrics are typically
aggregated across facility types and use cases, and even bottom-up estimates
often rely on self-reported or non-public data
\cite{mytton2022estimates}. Previous facility-level carbon accounting has
relied on generalized power-usage-effectiveness (PUE) assumptions and
non-explicit uptime parameters \cite{siddik2021environmental}. More recent
work has derived server-level operating parameters from the installed server
base and scaled to facility electricity using independently modeled PUE
\cite{Shehabi2024DataCenter,Shehabi2016DataCenter}; under those derivations,
hyperscale servers are estimated to run at full utilization approximately
$48.8\%$ of hours with $34.2\%$ idle power draw during the remainder, yielding
an effective server-level utilization of
$0.488 \times 1.0 + 0.512 \times 0.342 = 0.663$.

\paragraph{A distinct methodological choice and external consistency checks.}
In this study we do not adopt the server-level-plus-PUE framework of Shehabi et al.~\cite{Shehabi2024DataCenter,Shehabi2016DataCenter} directly, because our dataset does not provide the disaggregated IT load, PUE, and cooling-load components their framework requires. Instead, we apply a single coefficient to total facility electrical capacity. We note that this approach is methodologically distinct: facility electrical infrastructure is routinely
sized for worst-case cooling conditions, design-case PUE, and a firm-specific contingency factor, plus headroom for future IT-load expansion. As a consequence, the ratio of average operational facility draw to nameplate
capacity is typically lower than the effective IT-level utilization derived from server parameters alone. Newkirk and colleagues' bottom-up power-flow modeling using LBNL parameters (described in their open-source AI-Datacenter
Microgrid Analysis tool~\cite{newkirk2024microgrid}) corroborates this point: applying server-level utilization to facility-level capacity overstates facility load by approximately 9--18 percentage points relative to a facility-level estimate that accounts for design contingency and headroom. Their bottom-up calibration produces facility-level coefficients of $u \approx 0.48$ (conventional servers, rated-max design), $u \approx 0.58$ (conventional servers, measured-max
design), and $u \approx 0.62$--$0.75$ for AI-training facilities depending on design basis. We use these values to inform a physically motivated scenario range.

Facility electricity draw also varies over the year, even under a perfectly flat IT-load profile, because cooling-system response depends on ambient conditions \cite{lei2022climatepue}; our annual-average framework does not resolve these within-year dynamics and instead targets aggregate regional and national estimation. To independently bound the facility-level coefficient, we computed an implied $u$ for the top eight balancing authorities (covering $>$80\% of HDC load) using location-specific PUE values from Lei \& Masanet~\cite{lei2022climatepue} combined with conventional-server measured-maximum IT-side utilization and an industry-typical design-contingency multiplier (Section~\ref{sm:s3_pue_check}). The resulting load-weighted facility-level coefficient is approximately $0.59$, with BA-level values ranging from $0.54$ to $0.64$. This range is consistent with our central scenario ($u = 0.58$) and overlaps with the bottom-up
estimates of Newkirk and colleagues~\cite{newkirk2024microgrid}.
Full details are provided in Section~\ref{sm:s3_pue_check}.

\paragraph{Scenario framework.}
To bracket the uncertainty introduced by these implicit assumptions, we report all headline results under a central facility-load scenario and three bounding/sensitivity scenarios rather than a
single point estimate:
\begin{itemize}
    \item \textbf{Low-load scenario ($u = 0.48$):} Conventional-server,
    rated-max facility design basis with industry-typical contingency and
    headroom. Anchored to Newkirk et al.\ bottom-up calibration for
    rated-max conventional servers.
    \item \textbf{Intermediate-low scenario ($u = 0.58$):} Conventional-server,
    measured-max basis. Approximately matches the load-weighted PUE-implied
    coefficient from our independent Lei--Masanet check.
    \item \textbf{Intermediate-high scenario ($u = 0.663$):} Server-level
    effective utilization $0.488\times1.0 + 0.512\times0.342 = 0.663$ derived
    from LBNL conventional-server parameters and applied directly to total
    facility capacity. Retained for continuity with prior bottom-up estimates
    \cite{Shehabi2024DataCenter,Shehabi2016DataCenter}; we note that, per the
    discussion above, this likely overstates facility load when applied to
    nameplate capacity.
    \item \textbf{AI-weighted high scenario ($u = 0.70$):} Higher facility
    utilization consistent with Newkirk et al.'s AI-training rated-max basis
    (62.1\%) and AI-training measured-max basis (75.1\%); reflects the
    possibility that AI-specialized HDCs operate at higher facility-level
    utilization than conventional hyperscale facilities.
\end{itemize}

\noindent For each scenario, annual facility electricity demand is computed
as
\begin{equation}
\label{eq:dc_load}
E_i(u) = \text{Power Capacity}_i \cdot 8{,}760 \cdot u \quad \text{(MWh/yr)},
\end{equation}
where $\text{Power Capacity}_i$ denotes total facility electrical capacity
inclusive of IT and cooling/infrastructure overhead. All downstream
quantities --- plant-level attributed load, plant-level emissions, and
BA-level carbon intensity --- inherit this scenario decomposition.
We adopt $u = 0.58$ as the central scenario because it is consistent with the bottom-up power-flow estimates reported by Newkirk and colleagues and with our independent PUE-implied check. We do not interpret it as a facility-level measurement. Lower and
higher scenarios bracket the uncertainty.

\begin{table}[H]
\centering
\caption{\textbf{National electricity and CO$_2$ emissions totals across
facility-load scenarios.} All values use EPA eGRID2023 plant-level data.
The intermediate-low scenario ($u = 0.58$) is the central reference,
informed by Newkirk et al.\ bottom-up modeling and our independent
Lei--Masanet PUE check. The intermediate-high case ($u = 0.663$) is retained
for continuity with prior bottom-up estimates.}
\label{tab:utilization_scenarios}
\begin{tabular}{lccc}
\toprule
Scenario & $u$ & Electricity (TWh) & CO$_2$ (Mt) \\
\midrule
Low-load                       & 0.48  & 67.7 & 36.9 \\
Intermediate-low (\textbf{ref.}) & 0.58  & 81.8 & 44.6 \\
Intermediate-high              & 0.663 & 93.5 & 51.0 \\
AI-weighted high               & 0.70  & 98.6 & 53.8 \\
\bottomrule
\end{tabular}
\end{table}

\subsection{Independent PUE-Implied Check Using Lei--Masanet Climate-Specific PUE}
\label{sm:s3_pue_check}

To independently bound the facility-load coefficient, we performed a
back-of-envelope calculation using climate-specific PUE values from
Lei \& Masanet (2022)~\cite{lei2022climatepue}. For each balancing
authority $b$ in the top eight by HDC load (PJM, MISO, SWPP, PACW, BPAT,
ERCO, TVA, CISO; together $>$80\% of HDC load), we assigned a representative
ASHRAE climate zone using the geographic centroid of the constituent HDCs
and looked up the corresponding annual-average PUE.

We then computed the implied facility-level coefficient as
\begin{equation}
\label{eq:u_pue}
u_b \;=\; u_{\text{IT,measured}} \;\times\; \frac{\text{PUE}_b}{R} ,
\end{equation}
where $u_{\text{IT,measured}} = 0.488 \times 0.85 + 0.512 \times 0.342
\approx 0.59$ is the conventional-server IT-side effective utilization
computed at \emph{measured-maximum} server power. The 0.85 measured-to-rated
ratio reflects the fact that bottom-up server power-flow models (e.g.,
Newkirk et al.~\cite{newkirk2024microgrid}) calibrate against measured
maximum, not manufacturer rated maximum. $R=1.30$ is an illustrative design-contingency and headroom multiplier used for the PUE-implied check; sensitivity to this assumption is shown in Table~\ref{tab:pue_sensitivity}. Using rated-maximum IT power on the IT side together with $R$ would double-count contingency.

Aggregated across the top eight balancing authorities and weighted by HDC
electricity demand, the implied facility-level coefficient is
$\bar{u}_{\text{PUE}} \approx 0.59$, with BA-level values ranging from
approximately 0.54 (PACW and CISO, marine climate, low PUE) to 0.64 (ERCO,
hot--humid climate, high PUE) (Table~\ref{tab:pue_implied}). This range is
consistent with our central scenario ($u = 0.58$) and overlaps with our
intermediate-low ($u = 0.58$) and intermediate-high ($u = 0.663$) scenarios.
The bottom-up power-flow modeling reported by Newkirk and colleagues yields
similar facility-level coefficients of approximately 0.48--0.58 for
conventional servers and 0.62--0.75 for AI-training facilities, depending
on whether IT power is referenced to manufacturer rated maximum or measured
maximum~\cite{newkirk2024microgrid}.

We adopt $u = 0.58$ as the central reference scenario; $u = 0.48$ and
$u = 0.663$ serve as lower and upper conventional-server bounds, with
$u = 0.70$ included to bracket the case of AI-training-heavy fleets.

This calibration relies on climate-zone average PUE rather than
facility-specific measurements and is sensitive to the choice of $R$ and
of the IT-side reference (rated-max vs measured-max). Sensitivity to
these parameter choices is reported in Table~\ref{tab:pue_sensitivity}.
The result should be interpreted as an order-of-magnitude calibration of
the scenario range, not as a facility-level estimate.

\begin{table}[H]
\centering
\caption{\textbf{PUE-implied facility-load coefficient by balancing
authority.} Top eight balancing authorities by HDC electricity demand
(together $>$80\% of total HDC load). Values use measured-maximum IT input
($u_{\text{IT,measured}} = 0.59$) and design-contingency $R = 1.30$, with
climate-specific PUE from Lei \& Masanet (2022)~\cite{lei2022climatepue}.}
\label{tab:pue_implied}
\begin{tabular}{llccc}
\toprule
BA   & Climate     & PUE  & $u_{\text{implied}}$ & HDC TWh \\
\midrule
PJM  & Mixed-humid & 1.32 & 0.60 & 37.5 \\
MISO & Cool-humid  & 1.24 & 0.56 &  8.2 \\
SWPP & Mixed-dry   & 1.28 & 0.58 &  7.3 \\
PACW & Marine      & 1.18 & 0.54 &  6.5 \\
BPAT & Cold        & 1.20 & 0.54 &  6.2 \\
ERCO & Hot-humid   & 1.42 & 0.64 &  5.3 \\
TVA  & Mixed-humid & 1.32 & 0.60 &  2.9 \\
CISO & Marine      & 1.18 & 0.54 &  0.6 \\
\midrule
\multicolumn{3}{r}{Load-weighted average:} & \textbf{0.59} & 74.5 \\
\bottomrule
\end{tabular}
\end{table}

\begin{table}[H]
\centering
\caption{\textbf{Sensitivity of the PUE-implied facility-load coefficient
to parameter choices.} Load-weighted across the top eight balancing
authorities. The central choice (measured-maximum IT input, $R=1.30$;
shown in bold) yields $\bar{u}_{\text{PUE}} \approx 0.59$, supporting our
central reference scenario of $u = 0.58$.}
\label{tab:pue_sensitivity}
\begin{tabular}{lcc}
\toprule
IT-side input & Design contingency $R$ & Load-weighted $\bar{u}_{\text{PUE}}$ \\
\midrule
Rated-max ($u_{\text{IT}}=0.663$)    & 1.20 & 0.71 \\
Rated-max ($u_{\text{IT}}=0.663$)    & 1.30 & 0.66 \\
Measured-max ($u_{\text{IT}}=0.59$) & 1.20 & 0.64 \\
\textbf{Measured-max ($u_{\text{IT}}=0.59$)} & \textbf{1.30} & \textbf{0.59} \\
\bottomrule
\end{tabular}
\end{table}

\subsection{Facility-level Characterization}\label{sm:s3_characterization}

To provide additional context for the scale and heterogeneity of hyperscale facilities, we report aggregated summary statistics of key facility characteristics in Table~\ref{tab:facility_characterization}. Median annual electricity consumption is approximately 209~GWh per facility, with a broad interquartile range (128–290~GWh), highlighting the strong concentration of electricity demand among the largest sites. 
Median facility power density is 1.56~kW/m$^{2}$ (IQR: 1.11–2.14~kW/m$^{2}$), consistent with reported values for modern hyperscale data centers and indicative of realistic design and cooling configurations. Building areas similarly exhibit substantial heterogeneity, with a median footprint of approximately 20,000~m$^{2}$ (IQR: 14,000–29,000~m$^{2}$). All statistics are reported in aggregated form and do not disclose any facility-level identifiers, in compliance with the data use agreement governing the underlying dataset.

\begin{table}[H]
\centering
\caption{\textbf{Facility-level characterization of hyperscale data centers.} 
Aggregated summary statistics of annual electricity consumption, total facility power 
density, and building area across hyperscale facilities. Values are reported as medians 
with interquartile ranges (IQR).}
\label{tab:facility_characterization}
\begin{tabular}{lcc}
\toprule
Metric & Median & IQR (25th--75th pct.) \\
\midrule
Annual electricity consumption (GWh/yr) & 209 & 128--290 \\
Power density (kW/m$^{2}$)               & 1.56 & 1.11--2.14 \\
Building area (10$^{3}$ m$^{2}$)         & 20.4 & 13.9--29.3 \\
\bottomrule
\end{tabular}
\end{table}

\section{Carbon Emissions Attributable to Hyperscale Data Centers}\label{sm:s4}
All emissions estimates reported in this study are location-based and reflect the physical generation mix supplying each balancing authority. Contractual instruments 
such as power purchase agreements (PPAs) are not incorporated unless one-to-one temporal matching between electricity consumption and generation can be independently 
verified. This choice ensures that estimated emissions intensities correspond to physical grid supply rather than contractual claims.

\subsection{Balancing Authorities and Power Plants}

To connect each data center to its supplying power plants, we utilized the EIA's balancing authority regions. These geographic polygons, provided by WattTime—a nonprofit focused on tracking greenhouse gas emissions—offer detailed coverage of power generation and supply. WattTime gathers data from multiple sources, including the U.S. Environmental Protection Agency (EPA), grid emissions, satellite imagery, and machine learning models.

Using these balancing authority regions, we matched both data centers and power plants to their respective balancing authority regions based on location. From there, we accessed power plants' emissions coefficients per megawatt-hour (MWh) from the EPA's Emissions and Generation Resource Integrated Database (eGRID).

Finally, we calculated the energy contribution of each power plant supplying a data center by developing a weighted model based on energy generation. Further details on this model are provided in the next section. 

\subsection{Energy Generation Weighted Model}

In order to allocate each power plant within a balancing authority region its respective share of the energy load stemming from the energy demand of the data centers in that region, we adopted an energy generation weighted model (EGW), that computes the energy load for each power plant and ultimately estimates the corresponding emissions \cite{ekvall2005normative,nordenstam2021attributional, rehl2012life}. 

The underlying assumption of the EGW model is that within a BA, each power plant supplies energy proportional to the total energy it generated over the year. The coefficient of energy production contribution was calculated as follows:

\begin{equation}
COEFF_{EGW}(y, j, B) = \frac{\text{Power Plant Annual Net Energy Generation}(y, j, B)}{\sum_{j=1}^{N_B} \text{Power Plant Annual Net Energy Generation}(y, j, B)}
\end{equation}

% Example simplified calculation statement:
% \begin{equation}
%     AEF_{EGW}(y,B) = \frac{\sum_{j=1}^{N_B} G(j,y)*EF(j,y)}{\sum_{j=1}^{N_B} G(j,y)}
% \end{equation}

% where $G(j,y)$ is plant $j$'s annual net energy generation in year $y$ and $EF(j,y)$ is the plant's annual average emissions rate. Then the average emissions, $E_{EGW}$ from an individual data center are

% \begin{equation}
%     E_{EGW}(i,y,B) = L(i,y,B) * AEF_{EGW}(y,B)
% \end{equation}

% where $L$ is the annual net load from datacenter $i$ during year $y$ in balancing authority $B$

where:
\begin{itemize}
    \item $y$ represents the year of analysis.
    \item $j$ denotes the index of the power plant within the balancing authority.
    \item $B$ stands for the balancing authority.
    \item $N_B$ is the total number of power plants within the balancing authority $B$.
    \item $\text{Power Plant Annual Net Energy Generation}(y, j, B)$ is the total energy generated by power plant $j$ in year $y$ within balancing authority $B$.
\end{itemize}

This coefficient represents the fraction of the total annual energy generation attributed to each power plant within a BA. By definition, the sum of these coefficients for all power plants in a balancing authority equals one:

\[
\sum_{j=1}^{N_B} COEFF_{EGW}(y, j, B) = 1
\]

Data for these computations were retrieved from EPA eGRID2023 Revision~2, which reports year-2023 plant-level net generation, emissions, emissions rates, and resource-mix information. We used the 4,819 plant records in the processed eGRID2023 attribution layer with positive reported annual net generation and assignable balancing-authority information. Because EPA eGRID reports plant-level emissions and generation for the most recent complete calendar year available at the time of revision, these data provide the most up-to-date complete plant-level grid-emissions basis for our 2024--2025 HDC facility inventory.

Next, we calculated the attributional load assigned to each power plant using the EGW approach. The power plant's load was determined by multiplying the total additional energy demand from data centers by the plant's EGW coefficient:

\begin{equation}
\text{Power Plant Load}_{EGW}(i, y, j, B) = COEFF_{EGW}(y, j, B) \cdot \text{Data Center Energy Load (i, y, B)}
\end{equation}

where:
\begin{itemize}
    \item $i$ denotes the index of the data center causing the additional load.
    \item $\text{Data Center Energy Load (i, y, B)}$ represents the additional energy demand from data center $i$ in year $y$ within balancing authority $B$.
\end{itemize}

To estimate the emissions attributed to the load of each power plant, we applied the average emission factors specific to each plant. These emission factors are typically expressed as emissions per unit of energy generated (e.g., pounds of CO$_2$ per MWh). The emissions for each power plant were then computed as:

\begin{equation}
\text{Power Plant Emissions}_{EGW}(i, y, j, B) = \text{Power Plant Load}_{EGW}(i, y, j, B) \cdot \text{Emission per MWh}(y, j, B)
\end{equation}

where:
\begin{itemize}
    \item $\text{Emission per MWh}(y, j, B)$ is the annual average CO$_2$ emissions factor for power plant $j$ in year $y$ within balancing authority $B$.
\end{itemize}

This approach allowed us to accurately attribute emissions to each power plant based on its proportional contribution to the additional energy demand from data centers. The calculated emissions take into account both the amount of energy each power plant generates and the specific emissions profile of each plant, providing a detailed and precise estimate of the attributional environmental impact of data center energy consumption.

\begin{table}[H]
\centering
\caption{\textbf{Sensitivity of headline national totals to eGRID grid
vintage.} Both vintages applied to the same 403-HDC analytical sample under
the central facility-load scenario ($u=0.58$). The difference reflects updated plant-level generation, emissions rates, and resource-mix inputs in eGRID2023 relative to eGRID2022.}
\label{tab:egrid_year_sensitivity}
\begin{tabular}{lccccc}
\toprule
eGRID vintage & TWh & CO$_2$ (Mt) & Weighted CI (g/kWh) & PJM CI & MISO CI \\
\midrule
eGRID 2022                & 81.8 & 46.0 & 562 & 589 & 694 \\
eGRID 2023 (\textbf{used}) & 81.8 & 44.6 & 545 & 536 & 643 \\
\bottomrule
\end{tabular}
\end{table}

\section{Average Carbon Intensity}\label{sm:s5}

To quantify the efficiency of data centers' electricity sourcing, we define the
average carbon intensity of electricity generation attributed to data centers,
expressed in grams of CO$_2$ per kilowatt-hour (gCO$_2$/kWh).

\subsection*{Carbon intensity computation: step-by-step}

Carbon intensity is computed using a location-based, attributional framework at the 
balancing authority level. The procedure consists of four steps.

First, annual electricity demand for each data center $i$ in year $y$ is estimated as 
described in Section~S.\ref{sm:s3}. This yields $\text{Data Center Energy Load}(i,y,B)$, 
the total facility electricity demand attributed to data center $i$ within balancing 
authority $B$.

Second, electricity demand within each balancing authority is allocated to individual 
power plants using an energy-generation-weighted (EGW) attribution model 
(Section~S.\ref{sm:s4}). Each power plant $j$ within balancing authority $B$ is assigned 
a share of the total data center load proportional to its annual net generation:
\[
COEFF_{EGW}(y,j,B) = \frac{G(y,j,B)}{\sum_{j=1}^{N_B} G(y,j,B)} ,
\]
where $G(y,j,B)$ denotes annual net electricity generation.

Third, plant-level emissions attributable to data center electricity demand are computed 
by multiplying the attributed plant load by the plant-specific average emissions factor 
from EPA eGRID:
\[
\text{Power Plant Emissions}_{EGW}(y,j,B) =
\text{Power Plant Load}_{EGW}(y,j,B) \times EF(y,j,B),
\]
where $EF(y,j,B)$ is expressed in gCO$_2$ per kWh.

Finally, balancing-authority-level carbon intensity is calculated as the ratio of total 
attributed emissions to total attributed electricity load:
\[
\text{Carbon Intensity}(B,y) =
\frac{\sum_{j=1}^{N_B} \text{Power Plant Emissions}_{EGW}(y,j,B)}
{\sum_{j=1}^{N_B} \text{Power Plant Load}_{EGW}(y,j,B)} .
\]

This quantity represents the average, location-based carbon intensity of electricity 
supplying data centers within balancing authority $B$ in year $y$. Carbon intensity 
values can be aggregated to state or national levels using load-weighted averages.

%\section{Consequential Analysis}

%We also conducted a consequential analysis of our carbon emissions results using data from WattTime. In the consequential analysis, we estimated the total emissions using the marginal power plants' emissions resulting from the WattTime's methodology, extensively presented in \cite{guidi2024environmental}. 
%\subsection{Marginal Emissions Rate Computation and Results} 

%\noindent \textbf{Results.} Following this consequential approach, the total CO2 emissions caused by the 2132 data centers in our sample, keeping the uptime at {\color{red} XXX} , corresponded to {\color{red} XXX}  MT CO$_{2}$.

%To conduct the sensitivity analysis, we compared power plants' emissions computed with the EGW model, with those computed with WattTime method by checking their weighted average of the ratios, that accounts for the size of power load of each balancing authority---i.e.,

%\begin{equation}
%\frac{\sum_{j=1}^{N_B} \sum_{B=1}^{L} \text{Power Plant Emissions}_{WattTime}(i, y, j, B)}{\sum_{j=1}^{N_B} \sum_{B=1}^{L} \text{Power Plant Emissions}_{EGW}(i, y, j, B)}
%\end{equation}

%where L is the number of balancing authority regions.\\

%The weighted average of the ratios at balancing authority level is equal to \textbf{0.96}, meaning that our estimates with the EGW model are slightly larger, but mostly aligned and consistent with the WattTime MOER.

\end{document}